\documentclass[11pt]{article}

\usepackage[final]{acl}
\usepackage{times}
\usepackage{latexsym}
\usepackage[ruled,vlined]{algorithm2e}
\usepackage{amsmath}
\usepackage{amssymb}
\usepackage{multirow}
\usepackage{bbm}
\usepackage[table]{xcolor}
\usepackage[T1]{fontenc}
\usepackage{caption}
\usepackage[utf8]{inputenc}
\usepackage{microtype}
\usepackage{inconsolata}
\usepackage{graphicx}

\title{ToST: A Tree-of-Thought Socratic Teaching Framework for Multi-Path Guidance and Parallel Thinking}

\author{Feng Ling \\
  Beijing Normal University \And
  Heng Yu \\
  Beijing Normal University}

\begin{document}
\maketitle
\begin{abstract}
Large Language Models (LLMs) exhibit strong problem-solving abilities, positioning them as promising agents for Socratic teaching to guide students through step-by-step heuristic questioning. 
However, existing approaches typically adopt a one-problem–one-solution paradigm, restricting the teaching guidance to a single linear reasoning path. This design limits instructional flexibility, weakens error recovery, and restricts students' ability to engage in parallel thinking to explore multiple valid solutions.
To overcome these, we propose \textbf{ToST}, a \textbf{T}ree-\textbf{o}f-Thought \textbf{S}ocratic \textbf{T}eaching framework that explicitly supports multi-path guidance under a one-problem–multiple-solutions paradigm. ToST employs \emph{Parallel Sowing}, a parallel-thinking–oriented questioning strategy to encourage students to approach problems from diverse perspectives, and a \emph{Multi-Path Adaptive Guidance} mechanism to provide more robust and non-linear instructions across alternative solution trajectories.
Concurrently, to fill the void in systematically evaluating such non-linear instructional capabilities, we advance the task of multi-path Socratic guidance by establishing \textbf{MPSG-Bench}, a comprehensive benchmark that includes a dataset of 31K multi-path teaching dialogues and a five-dimensional evaluation framework grounded in the SOLO (Structure of Observed Learning Outcomes) theory to assess parallel-thinking guidance.
Experimental results demonstrate that ToST significantly enhances guidance success rates while empowering students to navigate and explore multiple solution paths more effectively under both automatic and human metrics.
\end{abstract}

\section{Introduction}

Large language models (LLMs) have become central to personalized AI tutoring~\cite{pedro2019artificial, zhai2021review}, with recent advancements focusing on expert pedagogical modeling~\cite{wang2024bridging} and Socratic questioning~\cite{zhang2024spl, liu2024socraticlm}. These methods aim to provide coherent, step-by-step scaffolding through instructional dialogues.

\begin{figure*}[t]
\centering
  \includegraphics[width=15.8cm]{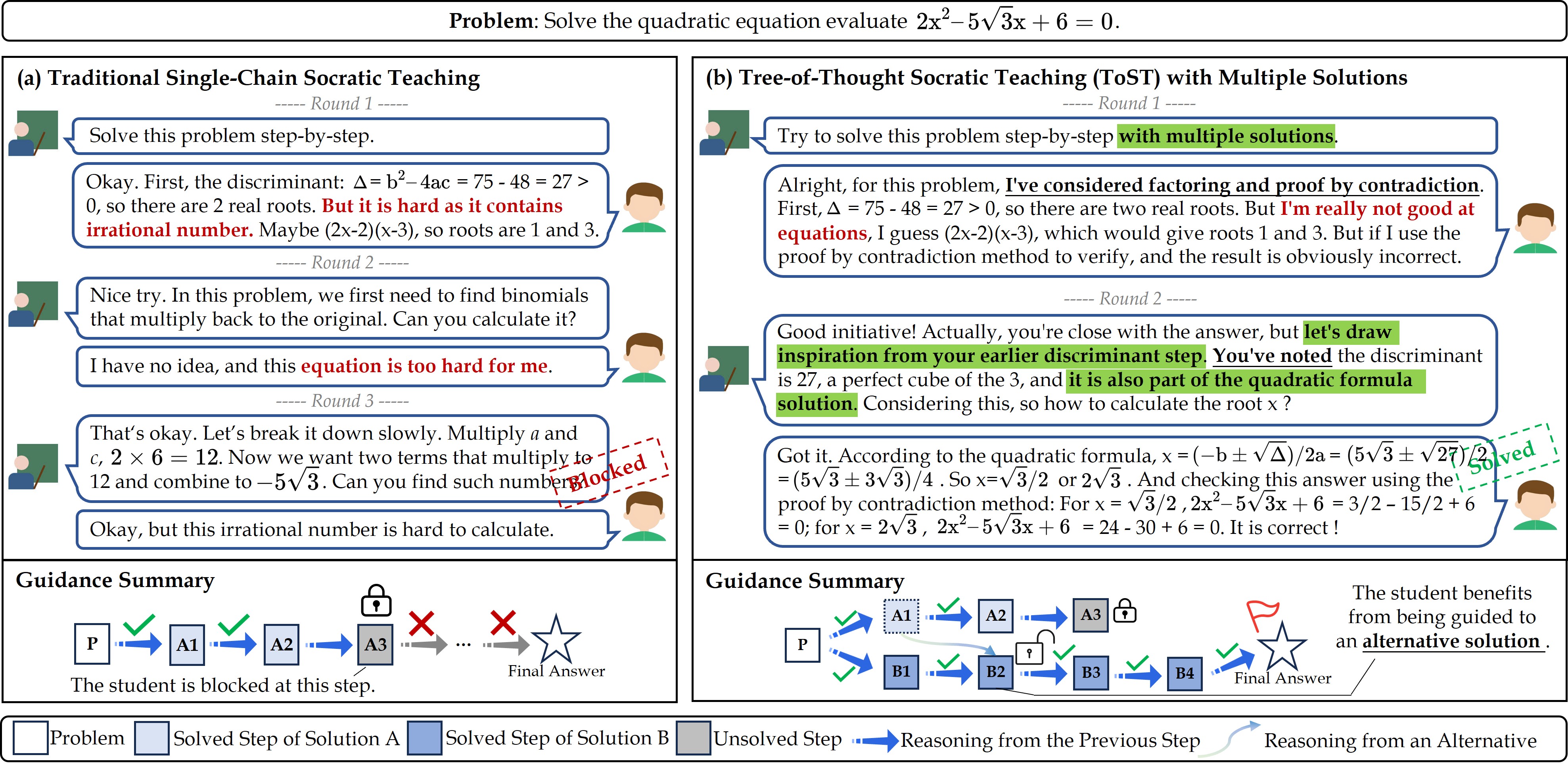}
  \caption{\textbf{Comparison between traditional Socratic Teaching and the proposed ToST paradigm.} Traditional method (left) employs a \textit{1P1S} paradigm with linear reasoning, susceptible to stalling from intermediate errors. In contrast, ToST (right) adopts a \textit{1PMS} paradigm via parallel thinking, enabling dynamic redirection when paths fail.} 
  \label{fig:motivate}
\end{figure*}

Despite these gains, most existing approaches are confined to a one-problem–one-solution (\textit{1P1S}) paradigm~\cite{gao2025agent4edu},
which restricts instructional flexibility and undermines the development of learners' parallel thinking \cite{pasztor2015technology}—an essential capability for navigating complex, non-linear problems from multiple perspectives~\cite{evans2014developing}.
As illustrated in Figure~\ref{fig:motivate}(a), traditional Socratic teaching typically anchors guidance to a single solution path, making it hard to recover from ineffective guidance~\cite{chu2025llm}.
These constraints motivate a shift toward a one-problem–multiple-solutions (\textit{1PMS}) paradigm that explicitly models multiple solution paths to provide more flexible and cognitively enriched guidance.

However, simply prompting LLMs to provide \textit{1PMS}-based guidance is challenging, since multi-path instruction imposes substantial System 2~\cite{kahneman2011thinking} planning demands: 
LLMs often struggle to track student progress across concurrent paths or decide when to switch trajectories. This difficulty is compounded by a lack of interpretable mechanisms for evaluating guidance strategies and estimating student cognitive states.

To address these challenges, we propose \textbf{ToST}, a Tree-of-Thought Socratic Teaching framework that 
instantiates the \textit{1PMS} paradigm via a \emph{parallel reasoning tree}.
ToST formalizes Socratic teaching as a tree-structured decision-making process, in which the parallel reasoning tree explicitly supports System~2 reasoning by tracking student progress across multiple concurrent solution paths and modeling pedagogically meaningful inter-path relationships.
As illustrated in Figure~\ref{fig:motivate}(b), ToST applies \emph{Parallel Sowing}, a questioning strategy that elicits multiple perspectives on the problem, followed by a multi-path adaptive guidance mechanism that dynamically decides whether to persist the current reasoning path or transition to alternatives when misconceptions arise.
During path transitions, teachers are encouraged to leverage \emph{reusable steps} (the shared steps between different paths) that the student has previously executed correctly.
These validated steps serve as familiar scaffolding to bridge transitions toward alternative strategies and encourage the exploration of new methods.

To fill the void in systematically evaluating non-linear instructional capabilities in \textit{1PMS} paradigm, we establish \textbf{MPSG-Bench}, a \textbf{M}ulti-\textbf{P}ath \textbf{S}ocratic \textbf{G}uidance benchmark comprising 31k teaching dialogues.
We propose a five-dimensional evaluation framework grounded in the Structure of Observed Learning Outcomes (SOLO) taxonomy~\cite{biggs2014evaluating}, enabling systematic assessment of teachers' \textit{1PMS} guidance and students' parallel-thinking ability. 
Experimental results on \textbf{MPSG-Bench} demonstrate that our ToST framework significantly improves guidance success rates by 11\% over educational LLM baselines and achieves more efficient exploration on multi-solution reasoning trees.

In summary, our contributions are:
\begin{itemize}
        \item We introduce the \textit{1PMS} paradigm to Socratic teaching and formalize it via a parallel reasoning tree, enabling explicit System~2 reasoning and inter-path pedagogical analysis to provide more adaptive instructional feedback.
        \item We propose \textbf{ToST}, a Tree-of-Thought Socratic Teaching framework to dynamically navigate concurrent solution paths,  improving instructional flexibility and robustness.
        \item We present \textbf{MPSG-Bench}, a new benchmark for multi-solution Socratic teaching, comprising 31k teaching dialogues and a SOLO-based evaluation framework to quantify structural complexity and parallel thinking across multiple solution paths, with extensive experiments validating the effectiveness of our approach.
\end{itemize}

\section{Related Work}

\subsection{LLMs-enhanced Personalize Tutoring}

Large language models (LLMs) have been widely adopted for personalized tutoring, supporting adaptive instruction~\cite{kasneci2023chatgpt,chen2024gptutor}, student simulation~\cite{zhan2025coderagent,gao2025agent4edu}, and learning path planning~\cite{gao2025graph,wang2025llm}.
Pedagogical alignment is commonly achieved through expert rules~\cite{wang2024bridging,park2024empowering}, educational pretraining~\cite{dan2024educhat}, or reinforcement learning~\cite{dinucu2025problem}, with Socratic prompting further enhancing personalization~\cite{liu2024socraticlm,chang2023prompting}.
However, most approaches rely on single-chain reasoning and adopt a one-problem–one-solution (1P1S) paradigm, limiting the cultivation of one-problem–multiple-solutions (1PMS) reasoning.

\subsection{Foundational Paradigms in Multi-Solution Pedagogical Reasoning}
\label{sec:solo}

The one-problem–multiple-solutions (\textit{1PMS}) paradigm promotes flexible problem-solving through the exploration of diverse valid solution paths~\cite{evans2014developing}.
This perspective aligns with parallel thinking, which promotes the simultaneous consideration of multiple reasoning strategies~\cite{de2016parallel}.
Learning progression in such settings is commonly assessed using the Structure of Observed Learning Outcomes (SOLO) taxonomy~\cite{biggs2014evaluating} (introduced in Appendix~\ref{sec:appendix_solo}), which provides a principled basis for our structure-aware, multi-path instructional framework.
Recent work extends linear chain-of-thought prompting~\cite{wei2022chain} with structured reasoning based on trees~\cite{yao2023tree}, graphs~\cite{besta2024graph}, or heuristic and value-based evaluation~\cite{browne2012survey}.
Parallel reasoning approaches aggregate independent solution chains via comparison or consensus~\cite{du2023improving,zheng2025parallel}, but lack explicit hierarchical representations and path-transition modeling.
In contrast, ToST explicitly constructs and navigates multiple solution trajectories within a Parallel Reasoning Tree, enabling process-aware multi-path instructional guidance.


\begin{figure}[t]
\centering
  \includegraphics[width=6cm]{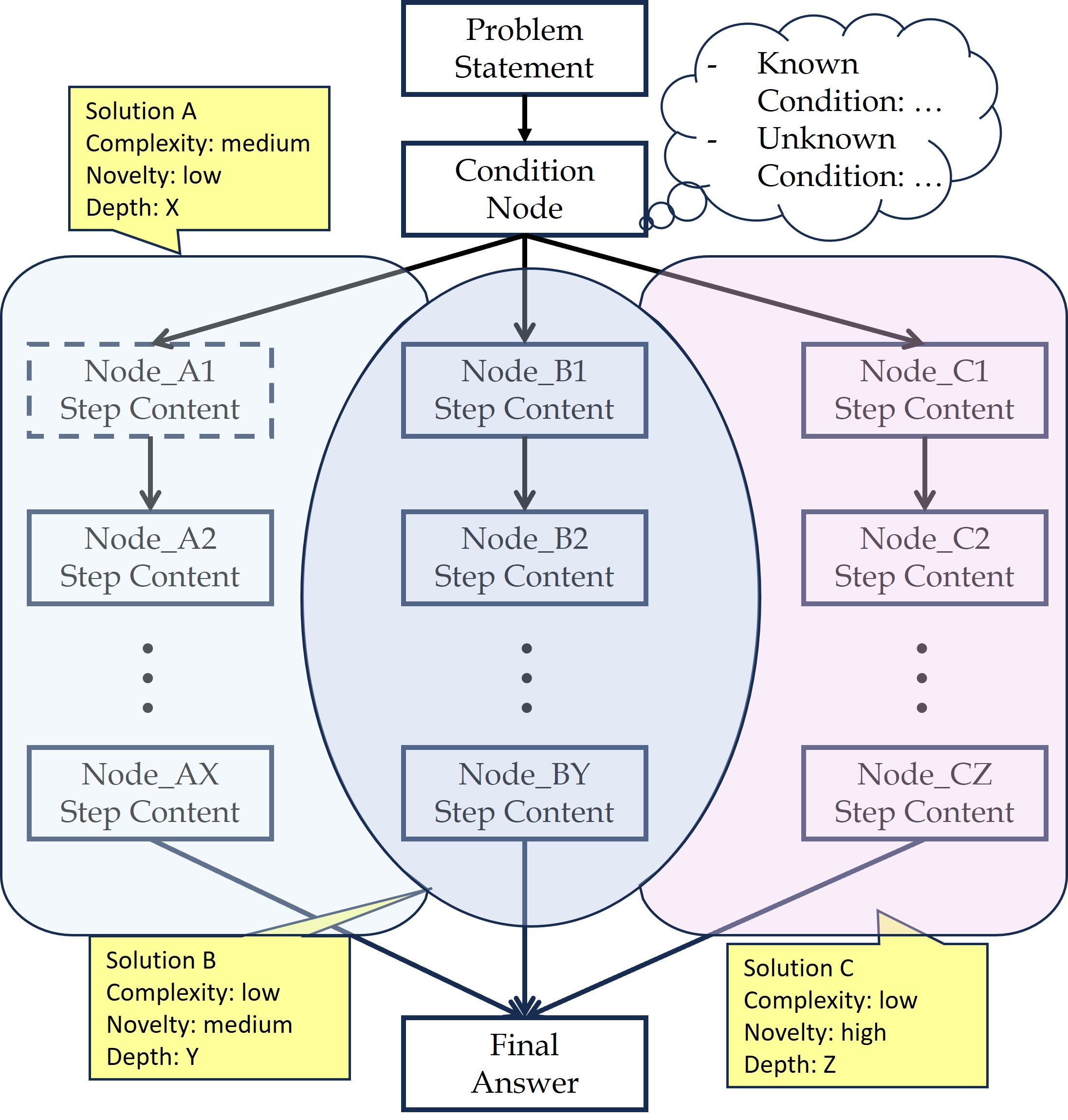}
  \caption{\textbf{Parallel Reasoning Tree (PRT) structure.} The root defines the problem, with branches representing alternative solution paths weighted by complexity, novelty, and depth. Nodes signify \emph{correct} intermediate steps leading to final answer.}
  \label{fig:tree}
\end{figure}

\begin{figure*}[t]
\centering
  \includegraphics[width=14cm]{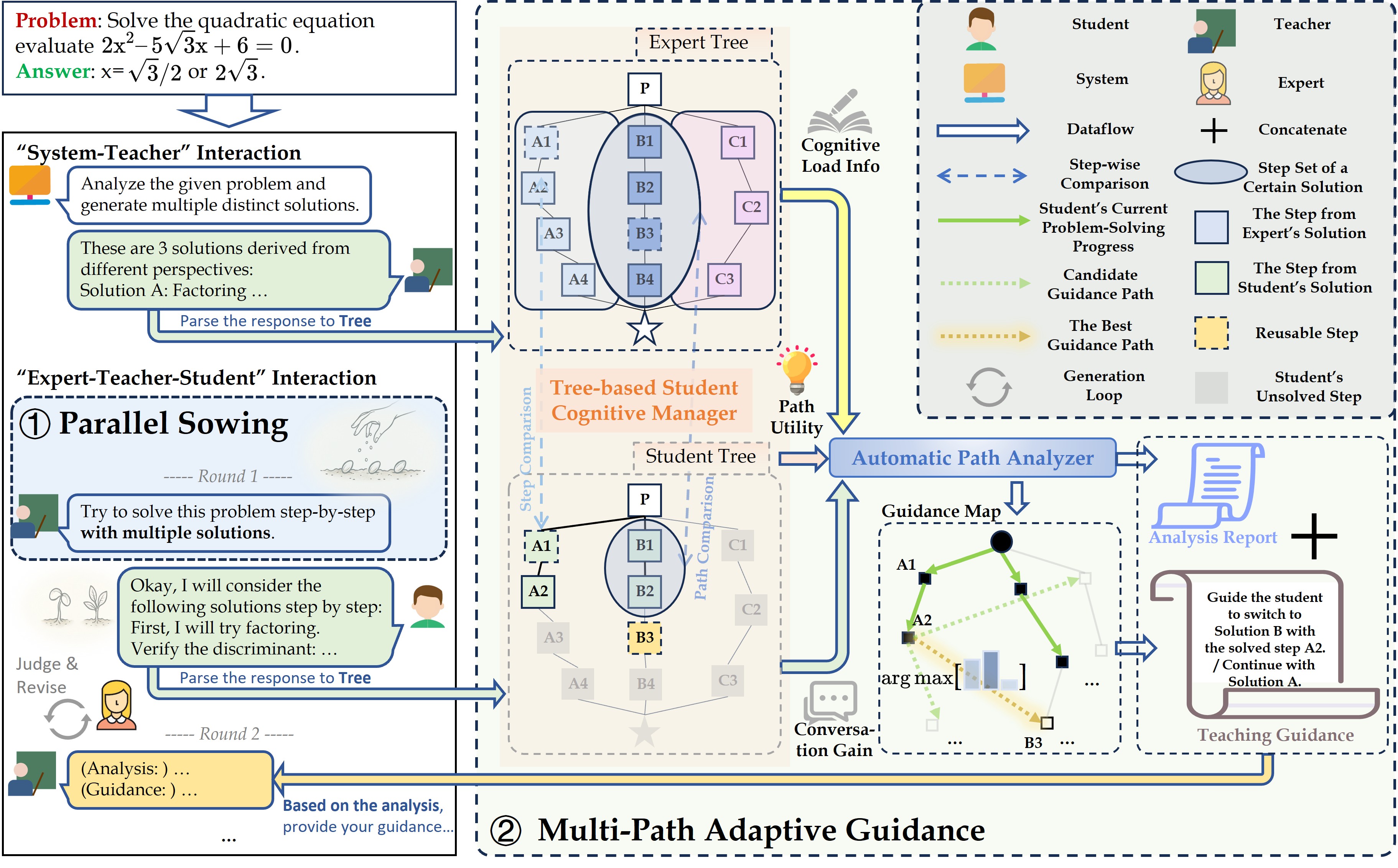}
  \caption{\textbf{Overview of the Tree-of-Thought Socratic Teaching (ToST) framework.} ToST operates on Parallel Reasoning Trees and consists of Parallel Sowing for multi-solution exploration and Multi-Path Adaptive Guidance for personalized Socratic instruction via student--expert tree comparison.}
  \label{fig:method}
\end{figure*}

\section{Background and Notation}
\label{sec:tree}

\subsection{Parallel Reasoning Tree Construction}
\label{sec:tree_construct}

The Parallel Reasoning Tree (PRT) represents multiple valid solution paths for a single problem within a unified hierarchical structure.
Each root-to-leaf path corresponds to a complete reasoning trajectory, explicitly representing alternative solution strategies in \textit{1PMS} settings. 
As illustrated in Figure~\ref{fig:tree}, each solution path is encoded as a complete reasoning trajectory, with explicit branching to capture divergence. 
Each path is annotated with three interpretable attributes: \emph{complexity}, reflecting intrinsic solution difficulty; \emph{innovativeness}, measuring structural deviation from canonical solutions provided by the official dataset; and \emph{depth}, defined as the number of intermediate reasoning nodes.
These attributes facilitate path-level comparisons and adaptive guidance.
The tree construction detail is introduced in Appendix~\ref{sec:appendix_data}, including the construction cost analysis in Appendix~\ref{sec:appendix_prt_cost}.

\subsection{Problem Definition}

Let $\mathcal{Q}$ denote a problem instance admitting a set of \emph{valid} solutions $\mathcal{S}=\{S_1,\dots,S_k\}$ with $k\geq2$. 
We formalize the Parallel Reasoning Tree (PRT) as a rooted directed acyclic graph $\mathcal{T}=(V,P,D,C,I)$, where $V$ comprises the root $v_0=\mathcal{Q}$, intermediate reasoning nodes, and solution leaves.
The set $P$ contains all root-to-leaf solution paths, and each path $p\in P$ is associated with a depth $D(p)\in\mathbb{N}$, complexity $C(p)\in\mathbb{R}$, and innovativeness $I(p)\in\mathbb{R}$.

Let $P_e\subseteq P$ denote expert-annotated paths and $P_s$ student-generated paths extracted from instructional interactions, where each student path corresponds to a (possibly partial) subpath of some $p\in P_e$.
Intermediate nodes in $P_s$ are partitioned into correctly solved nodes $K$ and unsolved nodes $M$, with a subset identified as reusable nodes $R\subseteq K\cup M$.
Given $\mathcal{T}$ and a student's partial trajectory $\tau$, the instructional objective is to select guidance actions that maximize learning effectiveness over $\mathcal{T}$ under the one-problem–multiple-solutions (1PMS) setting, by jointly optimizing student–expert path alignment and pedagogical factors such as reasoning depth and cognitive load.

\section{Tree-of-Thought Socratic Teaching}

As shown in Figure~\ref{fig:method}, ToST operates on an expert PRT $\mathcal{T}_e=(V_e,P_e,D_e,C_e,I_e)$ and an incrementally parsed student tree $\mathcal{T}_s=(V_s,P_s,D_s,C_s,I_s)$ to diagnose reasoning progress across concurrent solution paths.
It comprises \emph{Parallel Sowing}, which initiates multi-path exploration, and \emph{Multi-Path Adaptive Guidance} (MPAG), which performs tree-based diagnosis and guidance selection.

\subsection{Parallel Sowing}

Parallel Sowing initializes the student reasoning tree through teacher-guided multi-path exploration rather than by requiring the student to independently propose several complete solutions.
This step forms the initial branches of $\mathcal{T}_s$, establishing the structural basis for subsequent path-level diagnosis and guidance.
Importantly, Parallel Sowing does not assume that a novice learner can fully enumerate complete solutions.
The teacher may elicit partial intuitions, observable constraints, decomposition ideas, or candidate methods, each of which can serve as an entry point into a different branch of the PRT.

\subsection{Multi-Path Adaptive Guidance Strategy}

Multi-Path Adaptive Guidance (MPAG) provides path-aware instructional control over $\mathcal{T}_s$.
It comprises two modules: a \emph{Tree-based Student Cognitive Manager} for progress estimation and error localization, and an \emph{Automatic Path Analyzer} for guidance decision-making.

\subsubsection{Error Diagnosis with Tree-based Student Cognitive Manager}
\label{sec:diagnosis}

As dialogue proceeds, the Tree-based Student Cognitive Manager updates $\mathcal{T}_s$ and performs \textbf{PathMatch} for solution-category alignment and \textbf{NodeMatch} for intermediate-step correctness.

For each student path $P_s^{(i)}$, let $V_{P_s^{(i)}}$ denote the set of nodes along that path.
A progress score is computed as
\begin{equation}
\label{eq:node_match_path_merged}
\text{Score}_p(i)=
\frac{\sum_{v \in V_{P_s^{(i)}}} w_d(v)\,\omega(v)\, s(v,\mathcal{T}_e)}
{\sum_{v \in V_{P_s^{(i)}}} w_d(v)\,\omega(v)}
\end{equation}
where $s(v,\mathcal{T}_e)\in\{0,1\}$ indicates node-level correctness, $w_d(v)$ prioritizes earlier steps, and $\omega(v)$ gives higher weight to reusable nodes:
\begin{equation}
\label{eq:omega_merged}
\omega(v)=
\begin{cases}
\gamma, & v \in R, \\
\sigma, & v \in K \setminus R, \\
\delta, & v \in M \setminus R,\; R \neq \emptyset, \\
1.0, & v \in M \setminus R,\; R = \emptyset,
\end{cases}
\end{equation}
Here, $R$ denotes reusable nodes, $K$ correctly solved nodes, and $M$ unsolved nodes. 
The coefficients $\gamma$, $\sigma$, and $\delta$ are hyperparameters controlling the relative importance of reusable, correct, and incorrect nodes, respectively, with $\gamma \geq \sigma \geq \delta > 0$. 
This weighting favors stable and reusable reasoning steps, so the score reflects how much reliable structure can support subsequent guidance.

\subsubsection{Guidance Decision with Automatic Path Analyzer}

The Automatic Path Analyzer selects instructional paths by evaluating candidate expert paths.
Inspired by the global Product-of-Experts formulation~\cite{loula2025syntactic}, we model each instructional path as being constrained by multiple pedagogically relevant factors, each contributing a potential function to the overall guidance value.
Formally, for each candidate path $P_t \in P_e$, we define a guidance value $H(P_t)$ that integrates student progress, remaining problem difficulty, instructional investment, and cognitive load. 
The resulting guidance value is computed as
\begin{align}
\label{eq:value_func_merged}
H(P_t)=&
\underbrace{ \left[\alpha\,\text{Score}_p(t)+\frac{\beta}{1+E(P_t)}\right] \cdot \frac{I(P_t)}{C(P_t)} \nonumber  }_{\text{Path Utility Estimation}}\\& \cdot 
\underbrace{ \left(1+\rho\,\frac{T(P_t)}{T_{\max}}\right) }_{\text{Conversation Gain}}
\cdot 
\underbrace{ \frac{1}{1+\lambda D(P_t)}
}_{\text{Cognitive Load}}
\end{align}
where $E(P_t)$ is remaining difficulty, $I(P_t)$ and $C(P_t)$ are innovativeness and complexity, $T(P_t)/T_{\max}$ measures dialogue investment, and $D(P_t)$ is path depth.
Intuitively, $H(P_t)$ favors paths with reliable progress, manageable remaining difficulty, useful novelty relative to complexity, sufficient dialogue investment, and bounded cognitive load.

The next guidance path is selected using a greedy switching rule:
\begin{equation}
\label{eq:greedy_decision}
P_{\text{next}} =
\begin{cases}
P_{\text{c}}, \quad \text{if } H(P_{\text{c}}) + \theta_{\text{switch}} \geq H_{\max}^{-c}, \\[4pt]
\displaystyle\arg\max_{P \in P_e,\, P \neq P_c} H(P), \quad \text{otherwise}.
\end{cases}
\end{equation}
where $H_{\max}^{-c} = \max_{P \in P_e,\, P \neq P_c} H(P)$ denotes the highest guidance value among alternative paths excluding the current path $P_c$. 
The threshold $\theta_{\text{switch}} > 0$ controls the sensitivity of switching decisions.
It acts as an inertia margin: MPAG switches only when an alternative path is clearly better, which keeps the rule interpretable and reduces oscillation.

At each dialogue round, the Tree-based Student Cognitive Manager and the Automatic Path Analyzer jointly update $\mathcal{T}_s$ and select guidance actions, enabling adaptive \textit{1PMS} instruction to promote students' parallel thinking.

\section{Benchmark: MPSG-Bench}
\label{sec:benchmark}

Effective \textit{1PMS} instruction requires benchmarks that capture parallel reasoning beyond final-answer accuracy.
Existing educational datasets typically assume a single canonical solution, making them unsuitable for multi-path assessment.
To fill this gap, we introduce \textbf{Multi-Path Socratic Guidance Benchmark (MPSG-Bench)}: a large-scale benchmark of multi-path problem-solving trajectories and a SOLO-grounded evaluation framework for quantifying a model's ability of parallel-thinking guidance.

\subsection{Constructing MPSG-Bench with a Multi-Agent Pipeline}
\label{sec:mpsg_bench}
MPSG-Bench contains 31k PRT-grounded teaching dialogues with two components:

\paragraph{Enhanced Multi-Path Problem-Solving Collection.}
Seed problems from GSM8K~\cite{cobbe2021training} and MATH~\cite{hendrycks2021measuring} are expanded with DeepSeek v3.2 into annotated PRTs with approximately five distinct solution paths per problem.

\paragraph{Multi-Path Teaching Dialogue Dataset.}
Using these PRTs, \textit{Student}, \textit{Teacher}, and \textit{Expert} agents generate Socratic dialogues under six student archetypes (Appendix~\ref{sec:appendix_data}).

Together, these datasets facilitates the fine-tuning of LLMs on structured, non-linear instructional turns and provides a standardized testbed to quantify performance gaps in error recovery and path-consistency—capabilities that traditional single-path datasets are unable to measure.

\begin{table*}[t]
\centering
\caption{\textbf{Performance comparison.} Original student performance (before guidance) is reported as "Raw Students".
$\text{Acc}$ denotes final problem-solving accuracy, while $\text{TreeAcc}$, $\bar{N}_{\text{Round}}$, and $\text{TreeAcc-R}$ are defined in Section~\ref{sec:eval}.
$\bar{N}_{\text{Method}}$ denotes the average number of solution strategies attempted. \emph{ToST w/o PS} and \emph{ToST w/o MPAG} denote ablations removing Parallel Sowing, and Multi-Path Adaptive Guidance, respectively.}
\label{tab:performance}

\definecolor{lightgray}{gray}{0.92}
\definecolor{skyblue}{RGB}{240, 248, 255}

\resizebox{\linewidth}{!}{
\begin{tabular}{l|ccccc|ccccc}
\hline
\multirow{2}{*}{\textbf{Method}} 
& \multicolumn{5}{c}{\textbf{GSM8K}}
& \multicolumn{5}{c}{\textbf{MATH-500}}\\

\cline{2-11}
& Acc (\%)& TreeAcc & $\bar{N}_\text{Round}$  &$\text{TreeAcc-R}$& $\bar{N}_\text{Method}$ 
& Acc (\%)& TreeAcc & $\bar{N}_\text{Round}$  &$\text{TreeAcc-R}$& $\bar{N}_\text{Method}$ 
\\ 
\hline
\rowcolor{lightgray}
Raw Students& 48.49& 37.28 & ---  &---  & 1.04 & 61.20& 35.82 & ---  &---  & 1.16 \\

SocraticLM & 84.38& 57.47  & 3.33  &17.26 
& 1.95  & 92.60& 49.42  & 2.29  &21.58 
& 1.55  \\

\rowcolor{lightgray}
EduChat-R1-8b & 76.27
& 51.93& 3.94& 13.18 & 1.62
& 84.40& 44.39& 2.42& 18.34 &1.41\\

EduChat-R1-32b & 79.45& 54.37& 3.51& 15.49 & 1.78& 86.80& 47.18& 2.33& 20.25 &1.58\\

\rowcolor{lightgray}
TutorRL-7B & 89.61
& 60.17  & 2.86  &21.04 
& 1.91  & 91.80& 48.99  & 2.26  &21.68 
& 1.52  \\

DeepSeek V3.2 & 89.92
& 57.11  & 2.26  &25.27 
& 1.16  & 96.00& 50.13  & 1.89  &26.52 
& 1.53  \\

\rowcolor{lightgray}
GPT5& \textbf{98.56}
& 67.05&  1.97&34.04& 2.06& 98.20& 50.75 & 1.75 &29.00 
& 1.67 \\

\rowcolor{skyblue}
\textbf{ToST (Ours)} & \cellcolor{skyblue}98.18& \cellcolor{skyblue}\textbf{68.75}  & \cellcolor{skyblue}\textbf{1.68}  & \cellcolor{skyblue}\textbf{40.92} 
& \cellcolor{skyblue}\textbf{2.39}  & \cellcolor{skyblue}\textbf{99.20}& \cellcolor{skyblue}\textbf{56.01}  & \cellcolor{skyblue}\textbf{1.53}  & \cellcolor{skyblue}\textbf{36.61}
& \cellcolor{skyblue}\textbf{2.12}  \\

\rowcolor{lightgray}
ToST w/o PS & 96.36& 64.65& 1.76&36.73& 2.04  & 97.00& 50.68& 1.71&29.64 & 1.53\\

ToST w/o MPAG & 97.19& 66.12& 1.72& 39.02 & 2.23& 98.80& 54.88& 1.67& 32.86 &1.96\\

\rowcolor{lightgray}
Raw w/ PS& 51.40& 42.79& --- &---& 2.12& 65.00& 33.53& --- &---& 1.86\\
\hline

\multirow{2}{*}{\textbf{Method}} 
& \multicolumn{5}{c}{\textbf{AIME24}}
& \multicolumn{5}{c}{\textbf{AIME25}}\\
\cline{2-11}
& Acc (\%)& TreeAcc & $\bar{N}_\text{Round}$  &$\text{TreeAcc-R}$& $\bar{N}_\text{Method}$ 
& Acc (\%)& TreeAcc & $\bar{N}_\text{Round}$  &$\text{TreeAcc-R}$& $\bar{N}_\text{Method}$
\\
\hline
\rowcolor{lightgray}
Raw Students& 71.94& 46.08 & ---  &---  & 1.13& 31.25& 29.41 & ---  &---  & 1.07 \\

SocraticLM & 90.83& 55.70  & 2.25  &24.76 
& 1.32
& 63.13& 53.19  & 4.72  &11.27 
& 1.74  \\

\rowcolor{lightgray}
EduChat-R1-8b & 84.72& 49.72& 2.86& 17.38 
& 1.23
& 46.04& 30.15& 7.29& 4.14 &1.51\\

EduChat-R1-32b & 86.11& 51.68& 2.59& 19.95 & 1.29
& 53.33& 48.71& 5.43& 8.97 &1.64\\

\rowcolor{lightgray}
TutorRL-7B & 89.72
& 55.67  & 2.33  &23.89 
& 1.31
& 41.88
& 34.49& 6.72  &5.13 
& 1.53  \\

DeepSeek V3.2 & 96.39
& 59.49  & \textbf{1.65}  &36.05 & 1.31
& 72.50& 53.62  & 4.23 &12.68 & 1.49  \\

\rowcolor{lightgray}
GPT5& \textbf{97.78}
& 60.32&  1.74&34.67 & 1.29
& 73.13
& 47.71&  4.18&11.41 &1.45\\

\rowcolor{skyblue}
\textbf{ToST (Ours)} & \cellcolor{skyblue}96.67
& \cellcolor{skyblue}\textbf{63.39}  & \cellcolor{skyblue}\textbf{1.65}  & \cellcolor{skyblue}\textbf{38.42}& \cellcolor{skyblue}\textbf{1.73}
& \cellcolor{skyblue}\textbf{81.25}& \cellcolor{skyblue}\textbf{55.77} & \cellcolor{skyblue}\textbf{3.89} & \cellcolor{skyblue}\textbf{14.34} & \cellcolor{skyblue}\textbf{1.79} \\

\rowcolor{lightgray}
ToST w/o PS & 95.28
& 57.28&  1.82&31.47 
& 1.62
& 79.38
& 52.82&  4.03&13.11 
& 1.62\\

ToST w/o MPAG & 96.39& 61.74& 1.73& 35.69 & 1.71& 78.54& 54.19& 3.94& 13.75 &1.73\\

\rowcolor{lightgray}
Raw w/ PS& 76.11& 52.61& --- &---& 1.67& 35.21& 45.90& --- &---& 1.65\\
\hline
\end{tabular}
}
\end{table*}

\subsection{Parallel Thinking Evaluation System}
\label{sec:eval}

We propose a five-dimensional evaluation framework that instantiates the \textit{1PMS} objective via SOLO theory, to quantify structure-aware reasoning, path-sensitive diagnosis, and guidance efficiency.
Unlike prior methods that rely on final-answer accuracy or heuristic judgments~\cite{dinucu2025problem,liu2024socraticlm,yang2025llm}, our approach enables an interpretable, fine-grained assessment of a model's ability to facilitate parallel thinking.

\paragraph{(1) SOLO Advancement Score (SAS).}
SOLO Advancement Score (SAS) measures changes in students' cognitive structure before and after guidance, based on the SOLO taxonomy \cite{biggs2014evaluating}.
Student reasoning is independently assessed five times using GPT-5, and SAS is computed as the difference between post- and pre-guidance SOLO levels.

\paragraph{(2) Tree Accuracy  ($\text{TreeAcc}$).}
TreeAcc evaluates the correctness of students' internal reasoning structures within the PRT by jointly considering node-level reasoning quality, path-level methodological alignment, and final-answer correctness:
This approach is inspired by CodeBLEU~\cite{ren2020codebleu}, which combines abstract syntax trees (AST), data flow, and n-gram similarities to evaluate code generation tasks.
\begin{equation}
\label{eq:tree_acc}
\begin{aligned}
\text{TreeAcc} =
&\alpha_t \cdot \text{NodeAcc} + \beta_t \cdot \text{PathAcc} \\
&+ \gamma_t \cdot \text{FinalMatch}
\end{aligned} 
\end{equation}
where $\alpha_t+\beta_t+\gamma_t=1$, $\text{NodeAcc}=\frac{1}{|P_s|}\sum_{p \in P_s} \text{Score}_p(p)$, $\text{FinalMatch} =\mathbb{I}[\text{leaf}(p)\ \text{is correct}]$, $\text{PathAcc}$ measures alignment between student and expert solution paths, weighted by innovativeness and complexity, and $\bar{N}_\text{Round}$ denotes the average number of dialogue rounds.

Moreover, as the efficiency in \textit{1PMS} settings depends on both reasoning correctness and how quickly guidance facilitates convergence, we define \emph{Tree Accuracy per Round} ($\text{TreeAcc-R}$) as $\text{TreeAcc-R} =\frac{\text{TreeAcc}}{\bar{N}_\text{Round}}$ where $\bar{N}_{\text{Round}}$ is the average guidance round.
This metric reflects the accuracy–efficiency trade-off critical to \textit{1PMS} instruction.

\paragraph{(3) Diagnostic Precision (DP).}
DP evaluates the teacher's ability to identify misconceptions, recognize latent solution paths, and select appropriate interventions, assessed via LLM-based judgment.

\paragraph{(4) Persistence Gain (PG).}
PG measures average TreeAcc gains when guidance follows a student's current path. It evaluates single-path scaffolding and identifies the transition from prestructural to unistructural cognition.

\paragraph{(5) Switching Gain (SG).}
SG measures the average TreeAcc improvement following path switching.
It reflects the teacher's capacity to promote cross-method expansion and integration, which is critical for advancing students from unistructural to multistructural and relational levels.

Together, these metrics jointly quantify overall parallel thinking development (SAS, $\text{TreeAcc-R}$, DP) and stage-specific instructional effectiveness (PG, SG), enabling comprehensive and interpretable benchmarking of \textit{1PMS} teaching systems.

\section{Experiments and Results}
\label{sec:exp}

\subsection{Experimental Setup}
\begin{figure}
\centering
  \includegraphics[width=7.6cm]{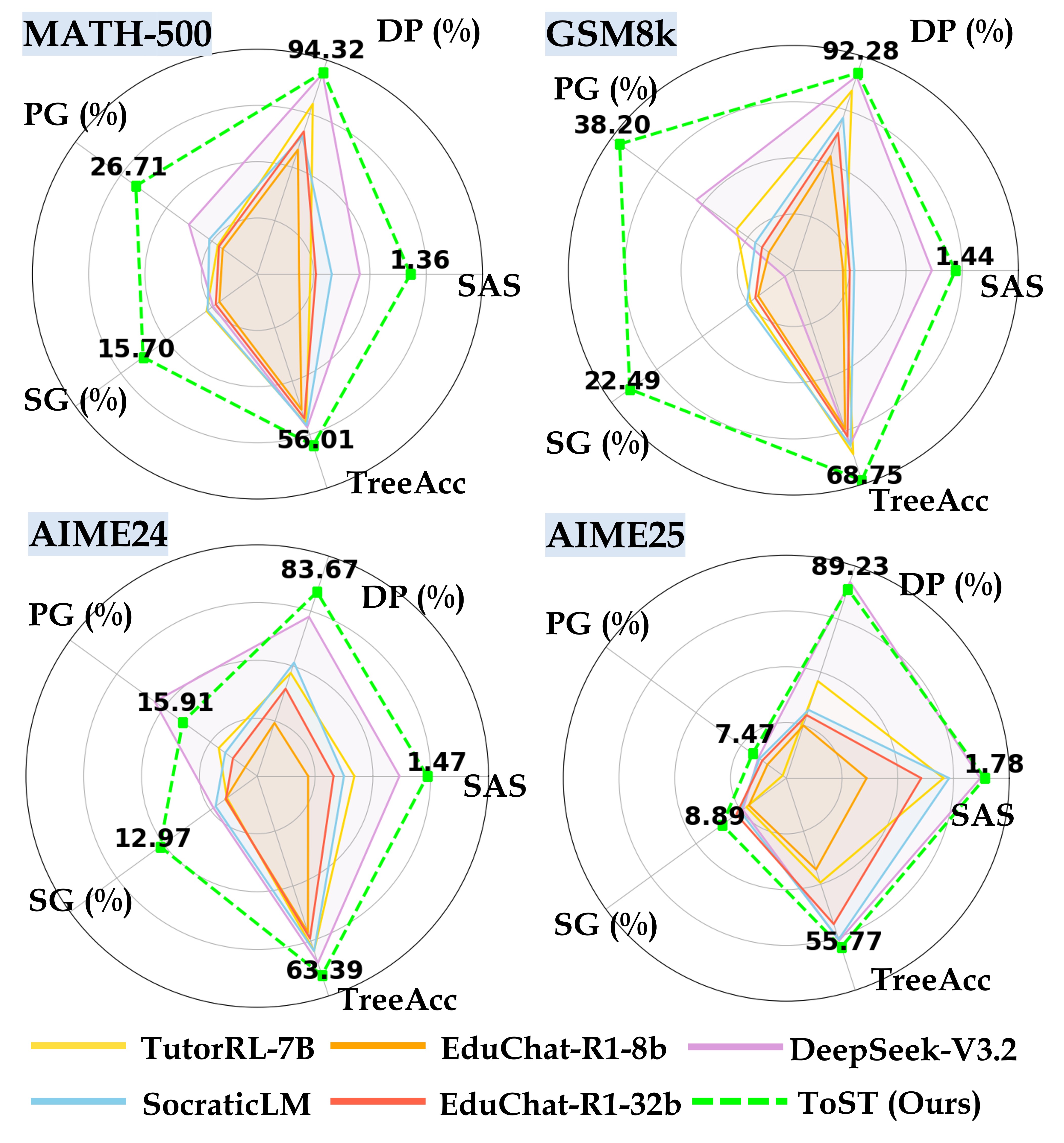}
\caption{\textbf{Comparison of guidance methods with five diagnostic protocol metrics} on GSM8k, MATH-500, AIME24, and AIME25.}
  \label{fig:eval_system}
\end{figure}

We compare ToST with educational agents (SocraticLM~\cite{liu2024socraticlm}, TutorRL-7B~\cite{dinucu2025problem}, EduChat~\cite{dan2024educhat}) and general LLMs (DeepSeek V3.2~\cite{deepseekai2025deepseekv3technicalreport}, GPT-5~\cite{openai2025gpt5}); for fair educational comparison, ToST uses \textsc{Qwen2.5-Math-7B-Instruct}~\cite{team2024qwen2} as the teacher model.
Experiments cover GSM8K~\cite{cobbe2021training}, MATH-500~\cite{hendrycks2021measuring}, AIME24, and AIME25~\cite{maa_aime_2024}; implementation and dataset details are in Appendices~\ref{sec:appendix_implementation} and~\ref{sec:appendix_data}.
Ablations remove Parallel Sowing (PS) via single-chain prompting or disable MPAG by removing explicit path selection; each experiment uses three random seeds.
Evaluation has two complementary parts: a large-scale simulated benchmark evaluation using MPSG-Bench protocol metrics (Table~\ref{tab:performance}, Figure~\ref{fig:eval_system}) and a small-scale human pilot evaluation in Section~\ref{sec:external_eval}.

\subsection{Main Results and Ablation Study}
\label{sec:exp_compare}

As shown in Table~\ref{tab:performance}, ToST consistently outperforms existing educational agents and strong general-purpose LLMs across all benchmarks.
Under the standard accuracy metric ($\text{Acc}$), ToST surpasses representative single-chain tutoring methods on every dataset, achieving an average improvement of 11\% in guidance success rate, while remaining competitive with state-of-the-art general LLMs despite using a substantially smaller 7B teacher model.
The advantage of ToST is more pronounced under $\text{TreeAcc-R}$, with gains of up to 20\% over the single-chain method \emph{SocraticLM} on GSM8K, underscoring the effectiveness of multi-path reasoning.
These improvements are consistent across problem difficulties; notably, on the challenging AIME25 benchmark, ToST outperforms the strongest general-purpose LLM baseline by approximately 8\%, indicating strong generalization to complex problem settings.
Ablations show that removing PS reduces explored strategies (e.g., 2.12 $\rightarrow$ 1.53 on MATH-500), while removing MPAG degrades solution quality; PS alone improves raw students by 3.71\% but is insufficient.
Figure~\ref{fig:eval_system} further shows consistent gains over educational baselines across all five internal protocol metrics, especially DP and SAS.

The PRT Parser reliability is confirmed by cross-model agreement analysis: average PDC and NMR reach 99.23\% and 97.69\%, respectively, across three LLM backends (see Appendix~\ref{sec:appendix_parse}).
Appendix~\ref{sec:appendix_prt_cost} quantifies the offline PRT construction overhead in terms of token use.
To clarify the intended domain scope of PRT, Appendix~\ref{sec:appendix_generalizability} provides case studies in physics, coding, and scientific reasoning.

\begin{table*}[t]
\centering
\caption{\textbf{Student study results (Track A, learner self-report).}
Participants rated their tutoring session on five dimensions using a 5-point Likert scale (mean $\pm$ std).
$\kappa$: Fleiss' inter-rater agreement.
Teacher/expert and LLM judge results are in Appendix~\ref{sec:appendix_human_eval}.
\textbf{Bold}: best per column.}
\label{tab:student_eval}
\definecolor{skyblue}{RGB}{240, 248, 255}
\resizebox{\linewidth}{!}{
\begin{tabular}{lcccccc}
\hline
\textbf{System} & \textbf{Helpfulness}$\uparrow$ & \textbf{Clarity}$\uparrow$ & \textbf{Understanding}$\uparrow$ & \textbf{Willingness}$\uparrow$ & \textbf{Burden}$\uparrow$ & $\kappa$ \\
\hline
SocraticLM     & 3.39$\pm$0.019 & 3.43$\pm$0.222 & 3.56$\pm$0.090 & 3.25$\pm$0.196 & 4.33$\pm$0.088 & 0.82 \\
EduChat-R1-32b & 3.21$\pm$0.078 & 3.02$\pm$0.062 & 3.65$\pm$0.065 & 3.14$\pm$0.097 & 4.19$\pm$0.078 & 0.75 \\
TutorRL-7B     & 3.75$\pm$0.002 & 4.02$\pm$0.211 & 4.86$\pm$0.011 & 3.73$\pm$0.216 & 4.37$\pm$0.066 & 0.80 \\
DeepSeek V3.2  & 4.42$\pm$0.117 & 4.13$\pm$0.119 & \textbf{4.92}$\pm$0.225 & 4.69$\pm$0.070 & 4.68$\pm$0.170 & 0.68 \\
\rowcolor{skyblue}
\textbf{ToST (Ours)} & \textbf{4.66}$\pm$0.128 & \textbf{4.68}$\pm$0.071 & 4.89$\pm$0.191 & \textbf{4.93}$\pm$0.209 & \textbf{4.95}$\pm$0.220 & 0.74 \\
\hline
\end{tabular}
}
\end{table*}

\subsection{Human Pilot Evaluation}
\label{sec:external_eval}

To further assess whether MPSG-Bench's structural advantages translate to human judgments and learner experience, we ran a human pilot study comparing blinded expert ratings, students' subjective responses, and an independent LLM-based judge.
Complete rubrics, prompts, interfaces, and additional results are provided in Appendix~\ref{sec:appendix_human_eval}.

\paragraph{Track A: Blinded human evaluation.}
We conduct two blinded small-scale studies on 100 dialogue sessions: an \emph{expert review} by \textbf{seven} mathematics teaching/tutoring raters and a \emph{pilot student study} with \textbf{ten} undergraduates interacting with all five systems in counterbalanced order.

\paragraph{Track B: Independent general LLM judge.}
As an auxiliary trend check, an independent general-purpose LLM judge rates 300 anonymized samples on helpfulness, clarity, diagnostic correctness, and switch naturalness.

\paragraph{Statistics and reporting.}
We report mean $\pm$ standard deviation, agreement statistics, and non-parametric tests, framing the small-scale study as pilot evidence. Table~\ref{tab:student_eval} presents learner-facing outcomes; teacher/expert and LLM-judge results are reported in Appendix~\ref{sec:appendix_human_eval}.

The learner self-reports further qualify the benchmark results.
ToST receives the strongest ratings on helpfulness, clarity, willingness to continue exploring, and perceived burden/manageability, indicating that learners found its structured multi-path guidance useful, understandable, and well paced.
This pattern is consistent with ToST's design: by preserving partial progress, seeding only a small number of teacher-selected alternatives when needed, and guiding transitions between them, the teacher can support exploration without making the interaction feel harder to manage.
The expert review and independent LLM judge in Appendix~\ref{sec:appendix_human_eval} provide the same broad signal: ToST ranks first or near first on the dimensions most directly tied to multi-path tutoring control.

\begin{figure}
\centering
\includegraphics[width=7.6cm]{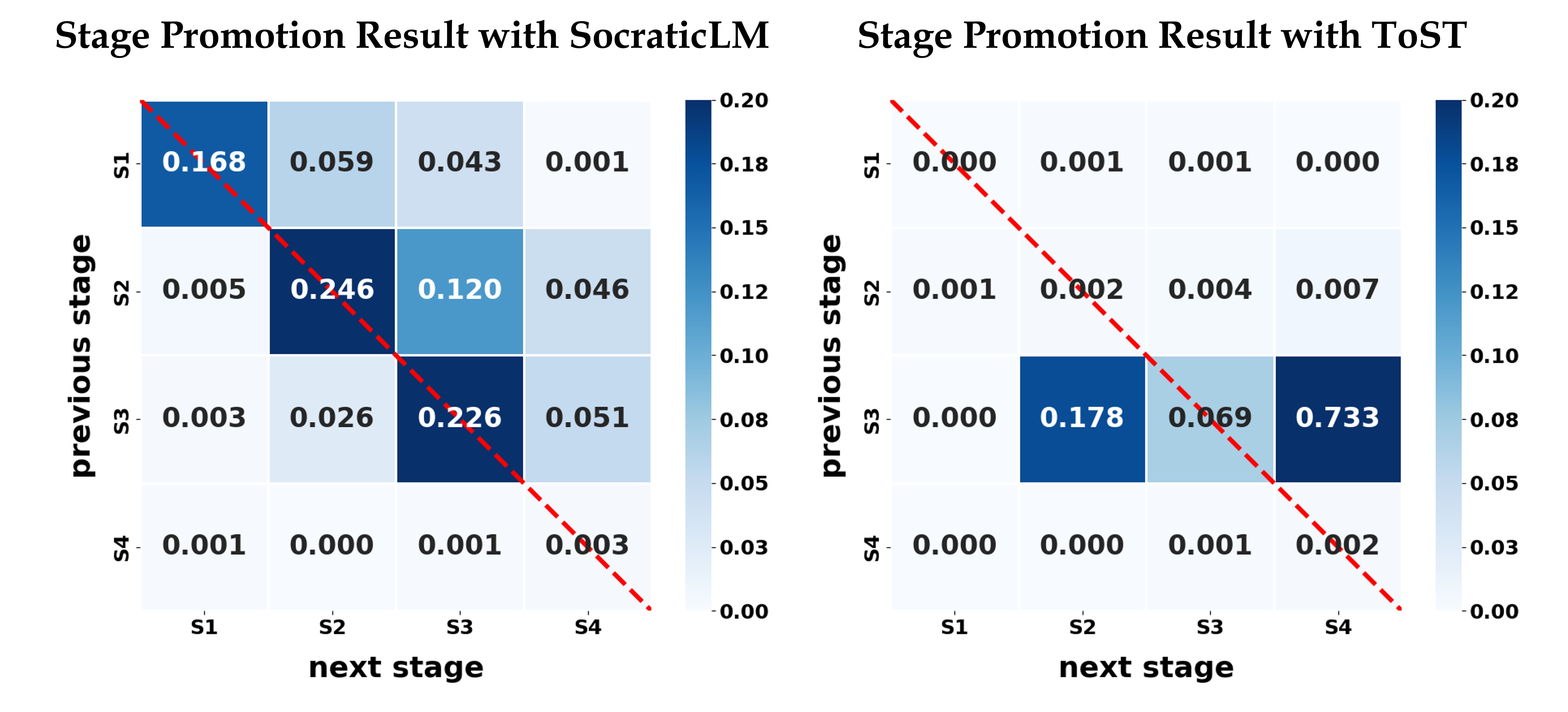}
\caption{\textbf{Stage promotion results with SocraticLM and ToST.} Diagonal entries indicate no cognitive progression; upper-right entries correspond to upward stage transitions. ToST induces substantially more upward transitions (e.g., 73.3\% S3$\rightarrow$S4); some downward entries (e.g., 17.8\% S3$\rightarrow$S2) reflect adaptive consolidation under increased task difficulty.}
\label{fig:eval_promote}
\end{figure}



\subsection{Pedagogical Advantages of the ToST Framework}

Figure~\ref{fig:eval_promote} presents a stage-transition analysis over multi-round instructional dialogues on all datasets, based on the SOLO taxonomy introduced in Section~\ref{sec:solo}.
Here, the \emph{previous stage} corresponds to the level of cognitive structure exhibited in students' initial problem-solving attempts, whereas the \emph{next stage} reflects the level attained following teacher guidance.
As shown, ToST consistently promotes upward transitions (e.g., 73.3\% S3$\rightarrow$S4), substantially exceeding SocraticLM. 
Compared to SocraticLM, Parallel Sowing in ToST elevates students' initial reasoning to higher SOLO levels.
In some cases, guidance appropriately consolidates multiple paths into a single focused strategy (e.g., 17.8\% S3$\rightarrow$S2), reflecting adaptive scaffolding under increased task difficulty.
These effects stem from ToST's explicit representation and management of reasoning diversity, which enables comparison across alternatives.

\section{Conclusion}

In this work, we propose ToST, a Tree-of-Thought Socratic Teaching framework that enables flexible, multi-path guidance under the one-problem–multiple-solutions paradigm. 
By integrating Parallel Sowing for diverse perspective exploration and Multi-Path Adaptive Guidance for robust instruction, ToST fosters deeper parallel thinking. 
We also introduce MPSG-Bench, a benchmark with 31K dialogues and a SOLO-based evaluation framework, to systematically assess multi-path teaching. 
Experiments show ToST significantly improves guidance effectiveness and supports richer exploration of solution paths.

\section*{Limitations}

While the ToST framework shows promising performance in guiding LLM-simulated students, several limitations exist.

Firstly, the current experiments and MPSG-Bench benchmark remain focused on mathematical problem-solving (e.g., arithmetic and algebra).
Secondly, ToST still depends on the availability of sufficiently reliable PRTs for the target problem domain.
Adaptive expansion of tree structure during tutoring is therefore an important direction for reducing upfront construction requirements.
Thirdly, ethical concerns, including potential over-reliance on AI-generated guidance and its impact on learners' autonomy and critical thinking, have not been systematically examined.

From a broader social and ethical perspective, improper deployment of ToST-like systems may risk encouraging passive learning behaviors or reinforcing narrow problem-solving patterns if instructional guidance is overly prescriptive.
Ensuring that such systems are used to support, rather than replace, human instruction and learners' independent reasoning is therefore essential.

Future research will focus on expanding the evaluation to larger cohorts and extending the MPSG-Bench dataset to include other domains. 
Additionally, efforts to mitigate bias and ensure ethical safeguards will be prioritized.

\section*{Ethical Considerations}

While LLMs offer educational potential, they also introduce risks regarding bias, data privacy, and the loss of human oversight. Moreover, their effects on learning outcomes must be continuously and rigorously evaluated to ensure pedagogical effectiveness and to prevent unintended negative consequences. To address these concerns and ensure pedagogical effectiveness, we developed a custom model that prioritizes granular data control. This approach prevents learner information from entering large-scale systems where re-identification risks exist, fostering a secure and trustworthy environment for our intelligent tutoring system.

\bibliography{custom}

\appendix

\section{Implementation Details}
\label{sec:appendix_implementation}

All experiments are implemented using PyTorch~2.8.0 with Python~3.12 on Ubuntu~22.04, and executed with CUDA~12.8 on two virtual GPUs, each equipped with 32~GB memory.
\textsc{Qwen2.5-Math-7B-Instruct}~\cite{team2024qwen2} is instruction-tuned~\cite{longpre2023flan} on the MPSG-Bench dataset as the ToST teacher using LoRA~\cite{hu2022lora}; detailed hyper-parameter settings are reported in Table~\ref{tab:hyper}.
Student behavior is simulated using GPT-3.5-turbo with six predefined student archetypes; one archetype is randomly sampled per session. 
Student responses are parsed into PRTs using DeepSeek V3.2.

To assess the effectiveness of ToST, we compare against representative educational tutoring models and strong general LLMs:

\paragraph{SocraticLM~\cite{liu2024socraticlm}.}
SocraticLM adopts a Socratic, thought-provoking instructional paradigm to deliver personalized guidance.
Its teacher model is trained on the SocraTeach dataset, which is constructed via a Dean--Teacher--Student multi-agent pipeline.
In our experiments, we evaluate the officially released checkpoint using the authors' recommended prompts.

\paragraph{TutorRL~\cite{dinucu2025problem}.}
TutorRL is trained via an online reinforcement learning framework that adapts LLMs into tutoring agents through expert-guided student--tutor simulations.
We employ the best-performing released checkpoint, which incorporates explicit reasoning tags to support pedagogically structured explanations of student errors.

\paragraph{EduChat~\cite{dan2024educhat}.}
EduChat is a large-scale educational chatbot guided by principles from cognitive psychology and learning sciences.
We evaluate both the 8B and 32B versions of the July~2025 released checkpoints, which introduce a ``thinking-before-teaching'' paradigm to enhance instructional coherence.

\paragraph{General LLMs.}
We additionally include DeepSeek~V3.2 and GPT-5 as strong general baselines.
Both models are prompted to act as tutors using standardized instructional prompts, without task-specific fine-tuning.
Evaluations are conducted via their official APIs using identical prompting protocols.

For all methods, the maximum number of interaction turns is capped at 10 to ensure fair comparison across models.
We use MARIO\_EVAL~\cite{zhang2024mario} to access the correctness of students' answer.

\begin{table}
\centering
\caption{Parameters for LoRA Finetuning (left) and ToST (right).}
\label{tab:hyper}
\resizebox{\linewidth}{!}{
\begin{tabular}{lc|cc}
\hline
\multicolumn{2}{c|}{\textbf{LoRA Fine-Tuning}} & \multicolumn{2}{c}{\textbf{ToST}} \\
\hline
\textbf{Parameter} & \textbf{Value} & \textbf{Parameter} & \textbf{Value} \\
\hline
         lora rank& 8 & $\gamma$ & 0.6\\
         lora alpha& 16 & $\sigma$ & 0.5\\
         lora target& q,k & $\delta$ & 0.1\\
         lora dropout& 0.1 & $\alpha$ & 0.52\\
         learning rate& 5e-5 & $\beta$ & 0.39\\
         weight decay& 0.1 & $\rho$ & 0.05\\
         train epochs& 3 & $\lambda$ & 0.16\\
         temperature& 0.95 & $\mu$ & 2\\
         length penalty& 1.0 & $\theta_\text{switch}$ & 0.2\\
         repetition penalty& 1.1 & $\alpha_t$ & 0.4\\
         trained context length& 6144 & $\beta_t$ & 0.2\\
         max new tokens& 1024 & $\gamma_t$ & 0.4\\
\hline
\end{tabular}
}
\end{table}

\section{MPSG-Bench Dataset: Construction and Statistics}
\label{sec:appendix_data}

This section describes the construction process and statistical characteristics of the proposed dataset within the Multi-Path Socratic Guidance Benchmark (MPSG-Bench). 
The dataset is designed to support \textit{1PMS} instruction and consists of two tightly coupled components: 
(i) an \emph{Enhanced Multi-Path Problem-Solving Collection} annotated with explicit Parallel Reasoning Trees (PRTs), and 
(ii) a \emph{Multi-Path Teaching Dialogue Dataset} built upon these structured problem representations.
Unless otherwise noted, GSM8k and MATH problems are drawn from their official releases, while AIME24 and AIME25 problems are collected from Zheng et al.~\cite{zheng2025parallel}.

\subsection{Enhanced Multi-Path Problem-Solving Collection}

\begin{figure}
\centering
\includegraphics[width=7.6cm]{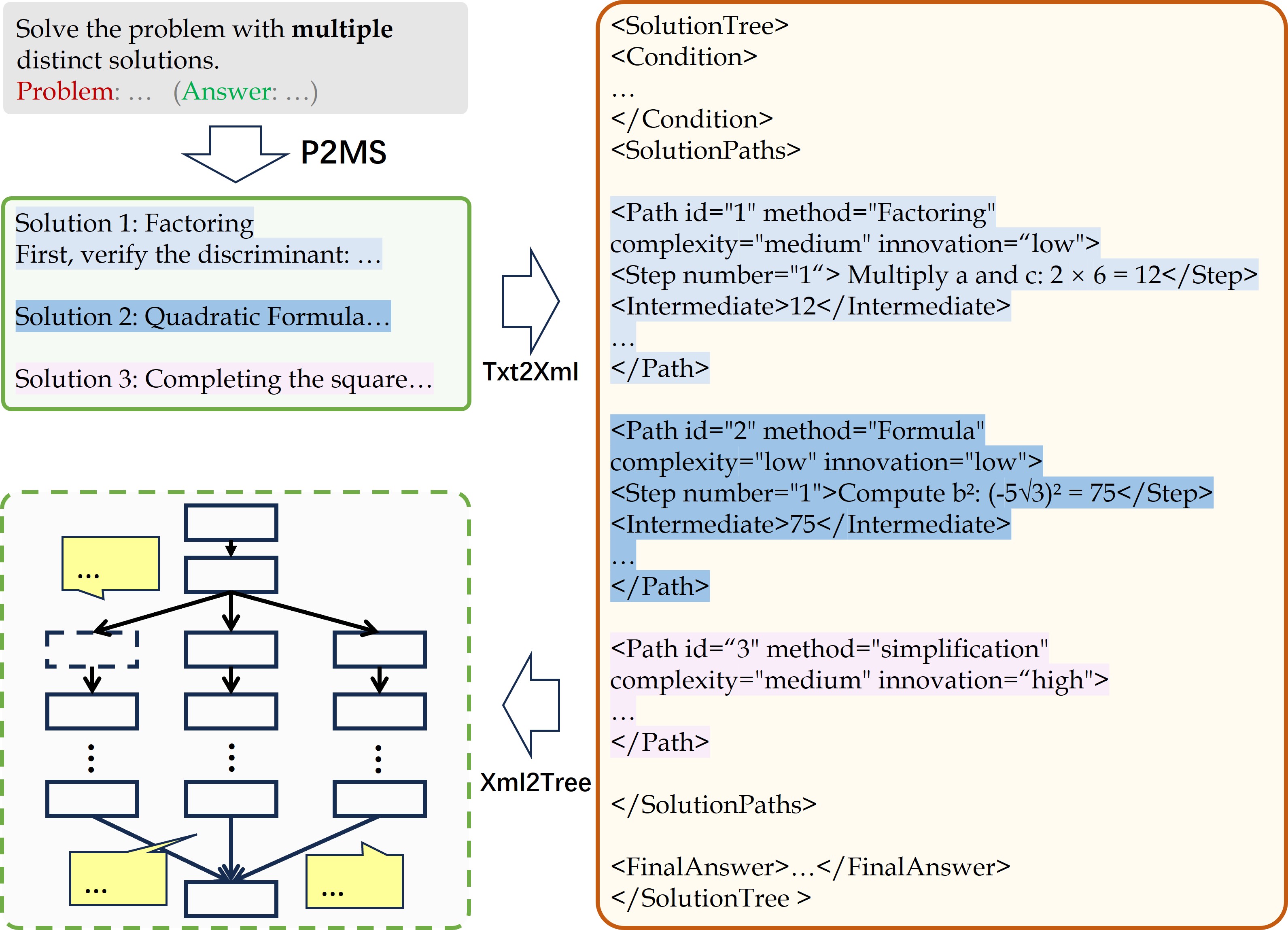}
\caption{\textbf{Overview of the Parallel Reasoning Tree (PRT) construction pipeline.} }
\label{fig:PRT_construct}
\end{figure}

The Enhanced Multi-Path Problem-Solving Collection augments existing mathematical problem-solving datasets with explicit PRT annotations.
As illustrated in Figure~\ref{fig:PRT_construct}, the construction of a PRT for each problem proceeds in three stages.
First, \emph{Problem-to-Multiple-Solutions} (P2MS) generates a natural-language response containing multiple methodologically distinct solution strategies.
Second, \emph{Text-to-XML} (Txt2Xml) extracts individual reasoning paths, annotates them with attributes such as \emph{complexity} and \emph{innovativeness}, and organizes them into a structured XML format.
Finally, \emph{XML-to-Tree} (Xml2Tree) parses the XML representation into a hierarchical and operational tree structure that explicitly captures parallel reasoning paths and their intermediate steps.

Table~\ref{tab:tree} reports statistics of the generated PRTs grouped by source dataset.
The results reveal a clear relationship between problem difficulty and tree structure.
Relatively simpler datasets (e.g., GSM8k and MATH) admit a broader range of viable solution strategies, resulting in a larger average number of reasoning paths, while their shorter solution chains yield fewer nodes per path.
In contrast, more challenging datasets (e.g., AIME25) constrain the solution space, producing fewer alternative paths but substantially deeper trees with a greater number of intermediate reasoning nodes.

\subsection{PRT Construction Cost Analysis}
\label{sec:appendix_prt_cost}

The PRTs used by MPSG-Bench are generated through the automated P2MS--Txt2Xml--Xml2Tree pipeline described above, and online tutoring only parses the learner's evolving response and applies guidance against the available problem-level structure.
Table~\ref{tab:prt_cost} reports the offline token cost of constructing problem-level PRTs by DeepSeek V3.2.

\begin{table}[h!]
\caption{\textbf{PRT construction cost analysis.}}
\label{tab:prt_cost}
\centering
\scriptsize
\resizebox{\linewidth}{!}{
\begin{tabular}{lccc}
\hline
\textbf{Dataset / Split} & \textbf{\# Problems} & \textbf{Avg. Input Tokens} & \textbf{Avg. Output Tokens} \\
\hline
GSM8k (Train) & 7{,}473 & 25{,}761 & 5{,}432 \\
MATH (Train) & 7{,}498 & 29{,}447 & 6{,}210 \\
GSM8k (Test) & 1{,}319 & 25{,}658 & 5{,}411 \\
MATH-500 & 500 & 32{,}369 & 6{,}826 \\
AIME24 & 360 & 30{,}292 & 6{,}388 \\
AIME25 & 480 & 36{,}786 & 7{,}757 \\
\hline
\textbf{Overall} & 17{,}630 & 27{,}901 & 5{,}884 \\
\hline
\end{tabular}
}
\end{table}

The results show that PRT construction is a bounded offline preprocessing cost rather than a per-turn tutoring cost.
Overall, each problem requires 27{,}901 input tokens and 5{,}884 output tokens on average.
The split-level variation mainly follows tree structure: AIME25 has the highest token cost because its PRTs are deeper, whereas GSM8k requires fewer tokens because its solution chains are shorter.
This suggests that reusable PRT resources are feasible to construct once for a problem set, while also motivating lighter two- or three-path trees and adaptive expansion for domains where full multi-path trees are unnecessarily costly.

\subsection{Multi-Path Teaching Dialogue Dataset}

Based on the constructed PRTs, we further build a Multi-Path Teaching Dialogue Dataset to support adaptive instructional guidance.
Following the multi-agent generation paradigm of SocraticLM~\cite{liu2024socraticlm}, we employ a pipeline consisting of \textit{Student}, \textit{Teacher}, and \textit{Expert} agents to generate approximately 31k instructional exchanges.
The \textit{Expert} agent refines teacher responses to ensure strict adherence to SOLO taxonomy constraints and enforces alignment with the guidance recommendations produced by the Automatic Path Analyzer in ToST.

Table~\ref{tab:dialogue} summarizes the statistics of the Multi-Path Teaching Dialogue Dataset.
The dataset is divided into a training set—comprising a Question-Scope-Oriented Single-Round Dialogue Corpus and a Student-Diversity-Oriented Multi-Round Dialogue Corpus—and a held-out Teaching Evaluation Set.

The \emph{Question-Scope-Oriented Single-Round Dialogue Corpus} focuses on increasing problem diversity.
For each problem drawn from GSM8k and MATH (difficulty levels 3--5), we generate a single-round instructional interaction that provides targeted guidance on the student's current reasoning state.
To prevent overfitting to specific learner behaviors, one of six predefined student cognitive archetypes is randomly assigned to each problem instance.
This design encourages broad coverage of problem types and solution structures while maintaining controlled interaction length, thereby promoting generalization across diverse problem distributions.

The \emph{Student-Diversity-Oriented Multi-Round Dialogue Corpus} is designed to increase diversity in student behaviors and instructional strategies.
This subset consists of multi-round dialogues constructed on MATH problems of lower difficulty (levels 1--2), which naturally admit extended interaction without excessive solution complexity.
For each problem, two distinct student cognitive archetypes are randomly sampled from the same pool of six, and guidance is delivered over multiple dialogue rounds.
This setting induces dynamic shifts in reasoning trajectories and instructional focus, enabling the teacher model to learn adaptive path continuation and path switching behaviors in response to evolving student states.

Together, these two corpora provide complementary supervision signals.
This design supports the training and evaluation of teacher models capable of delivering adaptive, personalized instruction across diverse student behaviors and reasoning trajectories.

\begin{table}[h!]
\caption{Statistics of the Enhanced Multi-Path Problem-Solving Collection in the MPSG-Bench dataset. Problems are grouped by their source datasets. For each subset, we report the number of problems, the average number of reasoning paths per PRT, and the average number of nodes per PRT.}
\label{tab:tree}
\centering
\scriptsize
\resizebox{\linewidth}{!}{
\begin{tabular}{lccc}
\hline
\textbf{Source Dataset} & \textbf{\# Problems} & \textbf{Avg. \# Paths} & \textbf{Avg. \# Nodes} \\
\hline
GSM8k (Train)  & 7{,}473 & 5.64 & 22.57 \\
MATH (Train)  & 7{,}498 & 4.95 & 25.80 \\
GSM8k (Test)   & 1{,}319 & 5.48 & 22.48 \\
MATH-500       &   500   & 4.90 & 28.36 \\
AIME24         &   360   & 4.76 & 26.54 \\
AIME25         &   480   & 4.17 & 32.23 \\
\hline
\textbf{Total} & 17{,}630 & 5.03 & 26.29 \\
\hline
\end{tabular}
}
\end{table}

\begin{table}[h!]
\caption{Statistics of the Multi-Path Teaching Dialogue Dataset in the MPSG-Bench dataset. We report the number of underlying problems and the total number of dialogue instances for each subset.}
\label{tab:dialogue}
\centering
\scriptsize
\resizebox{\linewidth}{!}{
\begin{tabular}{p{0.45\linewidth}cc}
\hline
\textbf{Subset} & \textbf{\# Problems} & \textbf{\# Dialogues} \\
\hline
Question-Scope-Oriented Single-Round Dialogue Corpus & 11.5k & 11.5k \\
Student-Diversity-Oriented Multi-Round Dialogue Corpus & 3.5k & 16.5k \\
Teaching Evaluation Set & 1.8k & 3.2k \\
\hline
\textbf{Total} & 16.8k & 31.2k \\
\hline
\end{tabular}
}
\end{table}

\subsection{Student Archetypes and Dialogue Diversity}

Student behaviors are simulated using six archetypes with distinct problem-solving preferences, including algebraic, logical, combinatorial, trial-and-error, equation-oriented, and non-specialized strategies.
These archetypes are grounded in Kolb's Learning Styles theory~\cite{kolb2014experiential}, which characterizes learners along two dimensions: abstract conceptualization versus concrete experience, and reflective observation versus active experimentation. While Kolb's framework defines four general learning styles, it remains coarse-grained for modeling domain-specific reasoning processes.

To better capture the diversity of mathematical problem-solving behaviors, we operationalize and extend these dimensions into six fine-grained archetypes. For example, algebraic and equation-oriented students emphasize abstract conceptualization, logical students exhibit reflective and structured reasoning, while trial-and-error and combinatorial students reflect stronger tendencies toward active experimentation. The non-specialized archetype represents learners without a dominant strategic preference.

To promote meaningful guidance interactions, student responses are repeatedly sampled (up to ten attempts) to increase the likelihood of incorrect initial answers. Dialogue diversity is further encouraged by combining single-round and multi-round interactions, randomly sampling student archetypes per problem, and balancing guidance that continues along the current reasoning path with guidance that explicitly triggers path switching. We adopt GPT-3.5-turbo to simulate students, as it exhibits relatively weaker problem-solving performance while maintaining strong role-playing and instruction-following capabilities.

\begin{table}[h]
\centering
\caption{\textbf{Consistency of solution tree parsing across APIs.} Agreement results on sampled MPSG-Bench student responses demonstrate reliability of PRT parsing.}
\label{tab:parse}
\resizebox{0.98\linewidth}{!}{
\begin{tabular}{lcc}
\hline
\textbf{Pairwise Comparison} & \textbf{PDC (\%)}
& \textbf{NMR (\%)} \\
\hline
DeepSeek V3.2 vs. GPT-5 & 99.48
& 98.78 \\
DeepSeek V3.2 vs. Gemini-2.5-pro & 99.68
& 96.92 \\
GPT-5 vs. Gemini-2.5-pro & 98.52
& 97.36 \\
\hline
Average & 99.23 & 97.69 \\
\hline
\end{tabular}
}
\end{table}

\begin{table}[h]
\centering
\caption{\textbf{Teacher/expert rubric for blinded dialogue review.} All dimensions use a 5-point Likert scale.}
\label{tab:expert_rubric}
\resizebox{\linewidth}{!}{
\begin{tabular}{p{0.30\linewidth}p{0.62\linewidth}}
\hline
\textbf{Dimension} & \textbf{Rating anchors (1 $\rightarrow$ 5)} \\
\hline
Diagnostic accuracy & Incorrect diagnosis or misses the student's actual issue $\rightarrow$ precisely identifies the student's misconception or latent valid path. \\
Guidance naturalness & Awkward, abrupt, or overly scripted tutoring language $\rightarrow$ smooth, conversational, and pedagogically natural tutoring language. \\
Preservation of correct progress & Discards or ignores correct partial reasoning $\rightarrow$ explicitly preserves and builds on what the student already got right. \\
Switch naturalness & Path transition is abrupt or confusing $\rightarrow$ transition is well-motivated and smoothly bridged through reusable intermediate steps. \\
Overall helpfulness & Guidance is unlikely to help the student progress $\rightarrow$ guidance is highly useful for moving the student forward. \\
Cognitive load appropriateness & Overwhelming or under-informative relative to student state $\rightarrow$ well-calibrated support with appropriate challenge and pacing. \\
\hline
\end{tabular}
}
\end{table}

\begin{table}[h]
\centering
\caption{\textbf{Student post-interaction questionnaire.} All dimensions use a 5-point Likert scale.}
\label{tab:student_rubric}
\resizebox{\linewidth}{!}{
\begin{tabular}{p{0.30\linewidth}p{0.62\linewidth}}
\hline
\textbf{Dimension} & \textbf{Rating anchors (1 $\rightarrow$ 5)} \\
\hline
Perceived helpfulness & The tutor was not helpful for solving the problem $\rightarrow$ the tutor was very helpful. \\
Clarity & The tutor's guidance was hard to understand $\rightarrow$ the tutor's guidance was very clear. \\
Understanding support & The interaction did not improve my understanding $\rightarrow$ the interaction substantially improved my understanding. \\
Willingness to continue exploring & I would not want to continue with this tutoring style in the future $\rightarrow$ I would like to continue exploring with this tutor. \\
Perceived burden & The interaction felt overly stressful or cognitively heavy $\rightarrow$ the interaction felt well-paced and manageable. \\
\hline
\end{tabular}
}
\end{table}

\begin{table*}[h]
\centering
\caption{\textbf{Teacher/expert blinded evaluation results (Track A).}
Expert raters evaluated anonymized tutoring dialogues from five systems on six pedagogical dimensions using a 5-point Likert scale (mean $\pm$ std).
$\kappa$: Fleiss' inter-rater agreement (substantial $>$0.6).
Student self-report results are in Table~\ref{tab:student_eval}.
\textbf{Bold}: best per column.}
\label{tab:expert_eval}
\definecolor{skyblue}{RGB}{240, 248, 255}
\resizebox{\linewidth}{!}{
\begin{tabular}{lccccccc}
\hline
\textbf{System} & \textbf{Diag.\ Acc.}$\uparrow$ & \textbf{Guidance Nat.}$\uparrow$ & \textbf{Pres.\ Correct}$\uparrow$ & \textbf{Switch Nat.}$\uparrow$ & \textbf{Overall Help.}$\uparrow$ & \textbf{Cog.\ Load}$\uparrow$ & $\kappa$ \\
\hline
SocraticLM     & 4.08$\pm$0.049 & 3.26$\pm$0.153 & 2.85$\pm$0.024 & 2.52$\pm$0.103 & 2.23$\pm$0.029 & 3.81$\pm$0.107 & 0.73 \\
EduChat-R1-32b & 4.35$\pm$0.148 & 3.51$\pm$0.174 & 3.69$\pm$0.022 & 3.37$\pm$0.216 & 2.34$\pm$0.010 & 3.37$\pm$0.244 & 0.76 \\
TutorRL-7B     & 4.80$\pm$0.071 & 3.17$\pm$0.001 & 3.96$\pm$0.238 & 4.15$\pm$0.228 & 3.35$\pm$0.231 & 3.66$\pm$0.046 & 0.74 \\
DeepSeek V3.2  & \textbf{4.96}$\pm$0.149 & 4.09$\pm$0.045 & 4.57$\pm$0.026 & 4.61$\pm$0.074 & 3.89$\pm$0.190 & \textbf{4.78}$\pm$0.211 & 0.70 \\
\rowcolor{skyblue}
\textbf{ToST (Ours)} & 4.94$\pm$0.090 & \textbf{4.19}$\pm$0.219 & \textbf{4.77}$\pm$0.203 & \textbf{4.80}$\pm$0.008 & \textbf{4.42}$\pm$0.218 & 4.66$\pm$0.181 & 0.75 \\
\hline
\end{tabular}
}
\end{table*}
\begin{table*}[h]
\centering
\caption{\textbf{Independent LLM judge results (Track B) and human--LLM agreement.}
An independent general-purpose LLM rated anonymized dialogues on four dimensions (mean $\pm$ std).
Win Rate: fraction of samples in which the system is preferred over all others.
$\rho_T$ / $\rho_S$: Spearman rank correlation with teacher / student scores.
\textbf{Bold}: best per column.}
\label{tab:llm_judge_eval}
\definecolor{skyblue}{RGB}{240, 248, 255}
\resizebox{\linewidth}{!}{
\begin{tabular}{lccccccc}
\hline
\textbf{System} & \textbf{Helpfulness}$\uparrow$ & \textbf{Clarity}$\uparrow$ & \textbf{Diag.\ Correct}$\uparrow$ & \textbf{Switch Nat.}$\uparrow$ & \textbf{Win Rate}$\uparrow$ & $\rho_T$$\uparrow$ & $\rho_S$$\uparrow$ \\
\hline
SocraticLM     & 3.45$\pm$0.307 & 3.78$\pm$0.232 & 3.11$\pm$0.136 & 2.80$\pm$0.827 & 0.34$\pm$0.031 & 0.82 & 0.82 \\
EduChat-R1-32b & 3.19$\pm$0.296 & 3.60$\pm$0.167 & 3.57$\pm$0.015 & 2.96$\pm$0.693 & 0.24$\pm$0.084 & 0.71 & 0.78 \\
TutorRL-7B     & 4.39$\pm$0.260 & 4.17$\pm$0.304 & 3.64$\pm$0.188 & 3.00$\pm$0.855 & 0.65$\pm$0.096 & \textbf{0.84} & 0.81 \\
DeepSeek V3.2  & 4.62$\pm$0.008 & 4.51$\pm$0.160 & 4.64$\pm$0.186 & 3.91$\pm$0.803 & 0.77$\pm$0.021 & 0.80 & 0.86 \\
\rowcolor{skyblue}
\textbf{ToST (Ours)} & \textbf{4.70}$\pm$0.107 & \textbf{4.69}$\pm$0.260 & \textbf{4.90}$\pm$0.119 & \textbf{4.17}$\pm$0.940 & \textbf{0.86}$\pm$0.066 & 0.74 & \textbf{0.87} \\
\hline
\end{tabular}
}
\end{table*}

\section{Cross-Domain Case Studies for PRT Generalizability}
\label{sec:appendix_generalizability}

This section provides conceptual case studies beyond mathematics to clarify the intended scope of PRT-based tutoring.
These examples are not benchmark-scale experiments; they illustrate when ToST can plausibly transfer: the task should admit explicit multi-step reasoning, multiple valid strategies or hypotheses, checkable intermediate states, and a shared target answer or evidence-based conclusion.

\paragraph{Physics.}
For an Atwood-machine problem with two masses connected over a frictionless pulley, a PRT can encode at least four valid paths: a system-level Newton's second-law solution, individual free-body equations, a work-energy derivation, and a Lagrangian derivation.
The root contains the shared givens ($m_1=5\,\mathrm{kg}$, $m_2=3\,\mathrm{kg}$, $g=9.8\,\mathrm{m/s^2}$), while leaves converge to the same acceleration, $2.45\,\mathrm{m/s^2}$.
Reusable nodes include quantities such as total mass, net force, acceleration direction, and shared energy terms.
Thus, if a learner stalls while solving simultaneous tension equations, MPAG can redirect the learner to the system-level force path while preserving correctly established physical quantities.

\paragraph{Coding.}
For computing the $n$-th Fibonacci number, a PRT can represent naive recursion, bottom-up iteration, memoized recursion, matrix exponentiation, and a closed-form formula.
Here, paths differ not by final value but by algorithmic structure, complexity, and implementation constraints.
Reusable nodes include base cases, recurrence relations, variable states, cache entries, and test cases.
Parallel Sowing can elicit whether the learner thinks recursively, iteratively, or through dynamic programming, while path switching can preserve correct base-case reasoning when guiding the learner from an inefficient recursive program to a more efficient iterative or memoized one.

\paragraph{Scientific reasoning.}
For a population-decline analysis, a PRT can organize competing causal hypotheses, such as habitat loss, invasive predators, disease, and climate-induced resource scarcity.
Each path begins from the same observation, e.g., a 60\% population drop over five years, and develops a different evidence chain using land-use maps, ship-arrival logs, pathology reports, or precipitation records.
The leaves need not be numeric answers; they can be evidence-weighted conclusions or a multi-factor synthesis.
Reusable nodes include correctly interpreted data sources and shared causal constraints, enabling the teacher to preserve valid observations while redirecting the learner toward a better-supported hypothesis.

Across these cases, the transferable component is not the mathematical content of MPSG-Bench, but the PRT abstraction: a structured set of alternative reasoning paths with verifiable intermediate states and reusable nodes.

\section{SOLO Taxonomy for Cognitive Structure Assessment}
\label{sec:appendix_solo}

The Structure of Observed Learning Outcomes taxonomy (SOLO)~\cite{biggs2014evaluating} is a hierarchical framework to assess the structural complexity of student understanding.
Rather than focusing on surface correctness, SOLO characterizes how well learners organize, integrate, and abstract knowledge.
It categorizes learning outcomes into five progressively sophisticated levels:

\paragraph{Prestructural (S1).}
The learner demonstrates little or no understanding of the problem, often responding with irrelevant or incorrect information.

\paragraph{Unistructural (S2).}
The learner identifies a single relevant aspect or solution strategy but applies it in a shallow or incomplete manner.

\paragraph{Multistructural (S3).}
The learner recognizes multiple relevant strategies or facts but treats them independently, without coherent integration.

\paragraph{Relational (S4).}
The learner integrates multiple solution paths into a coherent reasoning structure, demonstrating meaningful connections among ideas.

\paragraph{Extended Abstract (S5).}
The learner generalizes and abstracts beyond the immediate problem, applying principles at a theoretical or meta-cognitive level.

In this work, our evaluation focuses on the first four SOLO levels (S1--S4), as extended abstract reasoning (S5) is rarely exhibited in short-form tutoring dialogues and lies beyond the scope of the targeted student proficiency levels.

\section{Prompt Templates}
\label{sec:appendix_prompt}
\begin{figure*}
\centering
  \includegraphics[width=15.8cm]{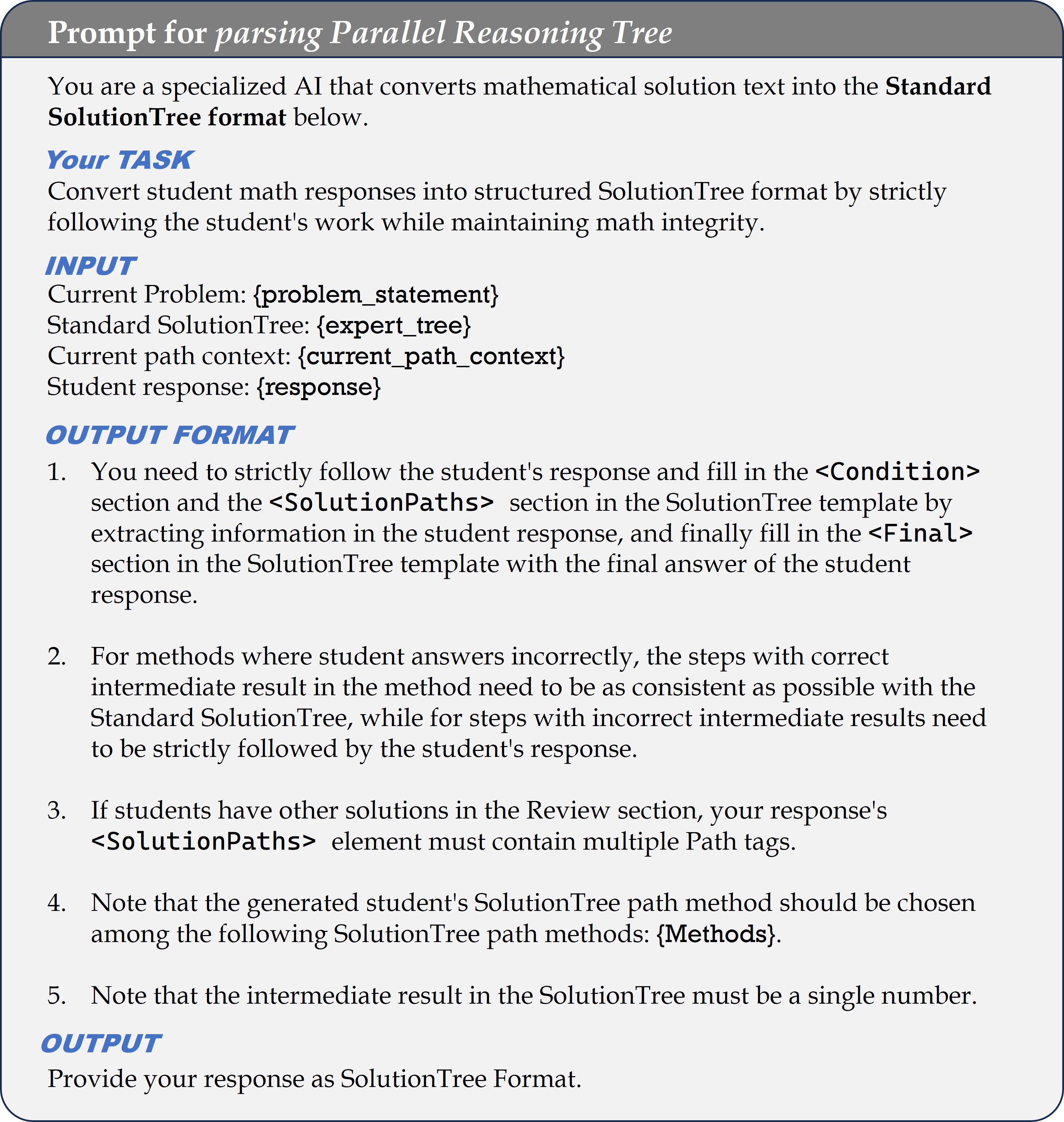}
  \caption{\textbf{Prompt template for Parallel Reasoning Tree parsing.}
  $\{\text{problem\_statement}\}$, $\{\text{expert\_tree}\}$, $\{\text{current\_path\_context}\}$, and $\{\text{response}\}$ denote placeholders for the target mathematical problem, the reference solution tree provided by the MPSG-Bench dataset (shown in the standard Solution Tree format), the student's current problem-solving trajectory, and the raw student response, respectively.}
  \label{fig:prompt_parse}
\end{figure*}

\begin{figure*}
\centering
  \includegraphics[width=15.8cm]{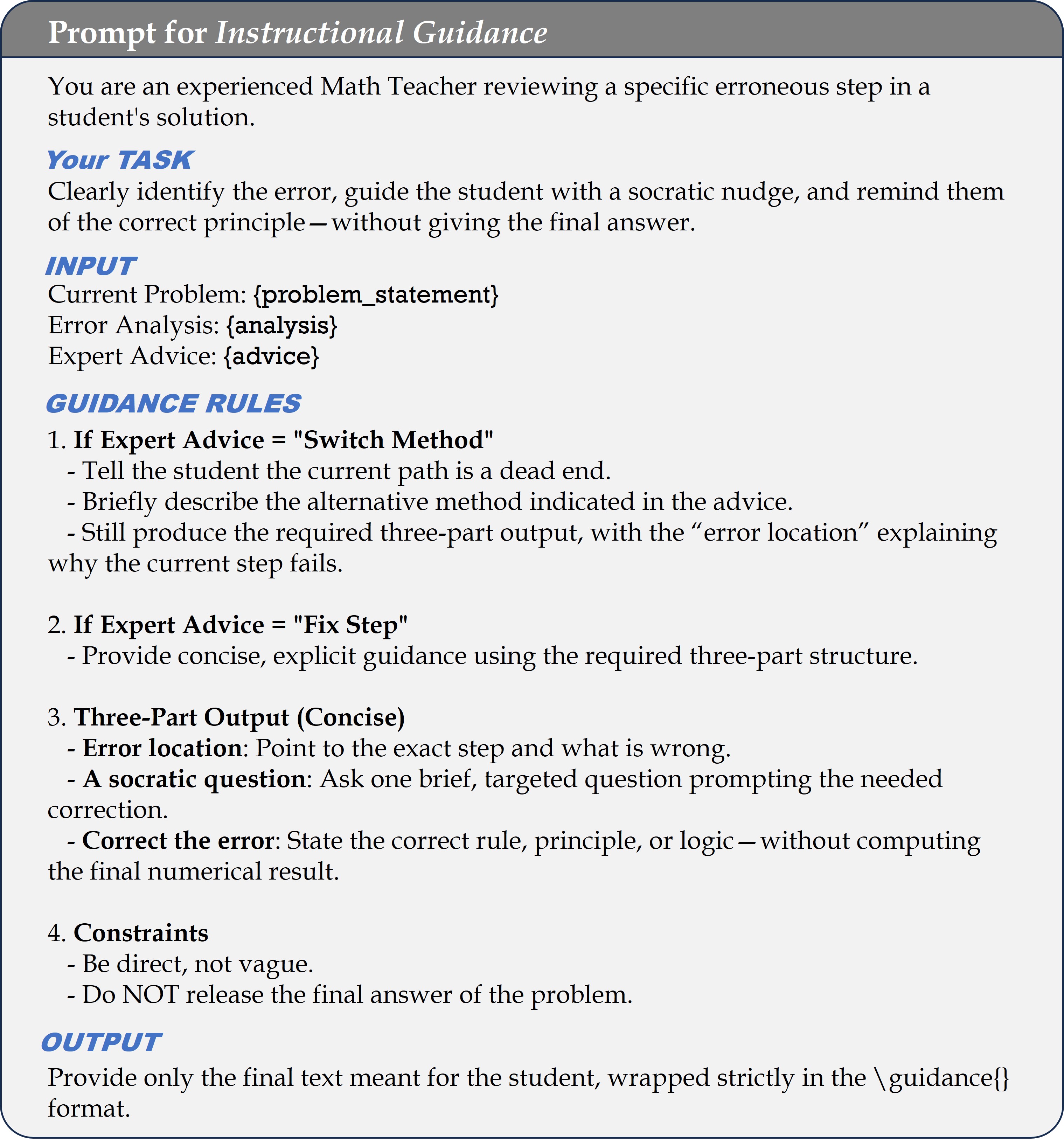}
  \caption{\textbf{Prompt template for instructional guidance under dual pedagogical modes.}
  $\{\text{problem\_statement}\}$, $\{\text{analysis}\}$, and $\{\text{advice}\}$ denote placeholders for the target problem, the path-level analysis report and the recommended instructional guidance provided by the Automatic Path Analyzer, respectively.}
  \label{fig:prompt_teach}
\end{figure*}

This section presents the prompt templates employed in ToST for (i) PRT parsing (Figure~\ref{fig:prompt_parse}) and (ii) instructional guidance under two modes: path continuation and path switching (Figure~\ref{fig:prompt_teach}).

\section{Human Pilot Evaluation Materials}
\label{sec:appendix_human_eval}

This section provides supplementary materials for the human pilot evaluation introduced in Section~\ref{sec:external_eval}, including rating rubrics, participant instructions, evaluation interface screenshots, and detailed full results for all tracks.

\subsection{Teacher/Expert Rubric}

Table~\ref{tab:expert_rubric} presents the full six-dimension rubric used in the blinded expert evaluation (Track A).
Each dimension is operationalized as a 5-point Likert scale with explicit behavioral anchors at both endpoints, ensuring that annotators apply consistent standards across dialogue samples from different systems.

\subsection{Student Rubric}

Table~\ref{tab:student_rubric} shows the five-dimension post-interaction questionnaire administered to student participants (Track A).
Items assess perceived helpfulness, clarity, understanding support, willingness to continue exploring, and cognitive burden, capturing both the effectiveness and the affective quality of the tutoring experience.

\subsection{Blinding, Annotation Protocol, and Statistics}

\noindent\textbf{Expert participants.} Seven graduate students and university instructors (4 graduate students in mathematics education or NLP, 3 university-level mathematics instructors) were recruited from top-tier universities. All had prior experience in mathematics tutoring or education research. Each expert independently rated 20 randomly sampled dialogue sessions per system (100 sessions in total).

\noindent\textbf{Student participants.} Ten undergraduate students (STEM majors) participated in the pilot student study. A counterbalanced within-subjects design was adopted: each participant completed one tutoring session with each of the five systems in a randomized order, with a five-minute break between sessions to reduce fatigue.

\noindent\textbf{Blinding.} System identities are hidden during annotation. Dialogue samples are randomized and presented as \texttt{System A}, \texttt{System B}, and (when needed) \texttt{System C}.
\textbf{Annotation unit.} Each unit contains the problem statement, optional expert reference solution, student dialogue history, and one tutoring continuation or full tutoring session depending on the study condition.
\textbf{Preference task.} When pairwise comparison is enabled, annotators select the more helpful tutoring response and may optionally provide a short free-text reason.
\textbf{Statistics.} Likert ratings are reported as mean $\pm$ standard deviation; pairwise preferences as win rate; and inter-rater agreement as Fleiss' $\kappa$ computed across expert raters. For the student study, $\kappa$ reflects cross-participant consistency in subjective ratings rather than annotation agreement, and should be interpreted accordingly.
\textbf{Ethics.} Participants receive a short consent form stating the study purpose, expected duration, anonymization policy, and the right to withdraw at any time.

%
%

\subsection{Evaluation Interface Screenshots}

\begin{figure*}[h]
\centering
\includegraphics[width=14cm]{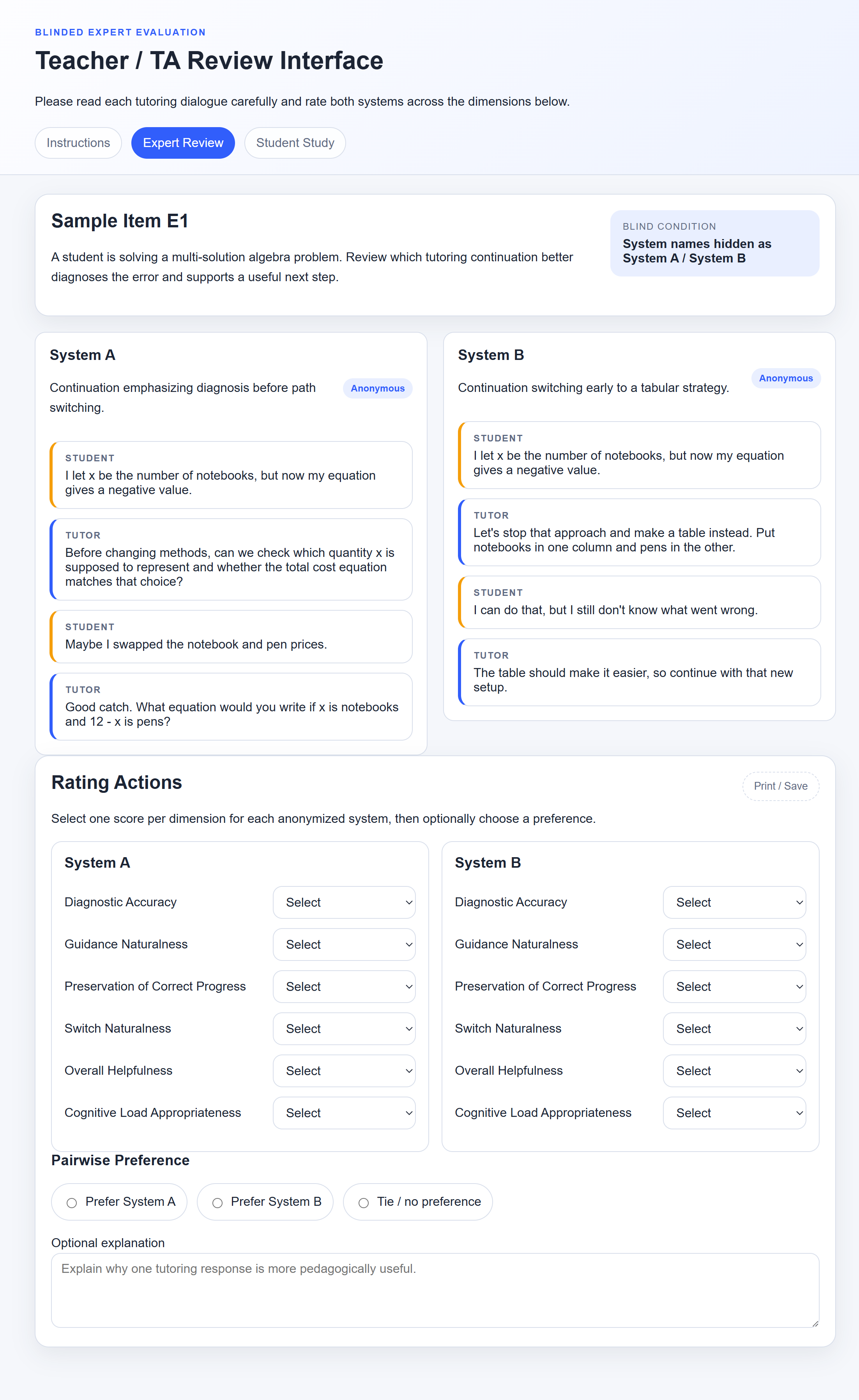}
\caption{\textbf{Expert-review interface.} The blinded rating interface presents anonymized tutoring dialogues side-by-side and collects six-dimension Likert scores with optional free-text rationale.}
\label{fig:expert_interface_placeholder}
\end{figure*}

\begin{figure*}[h]
\centering
\includegraphics[width=14cm]{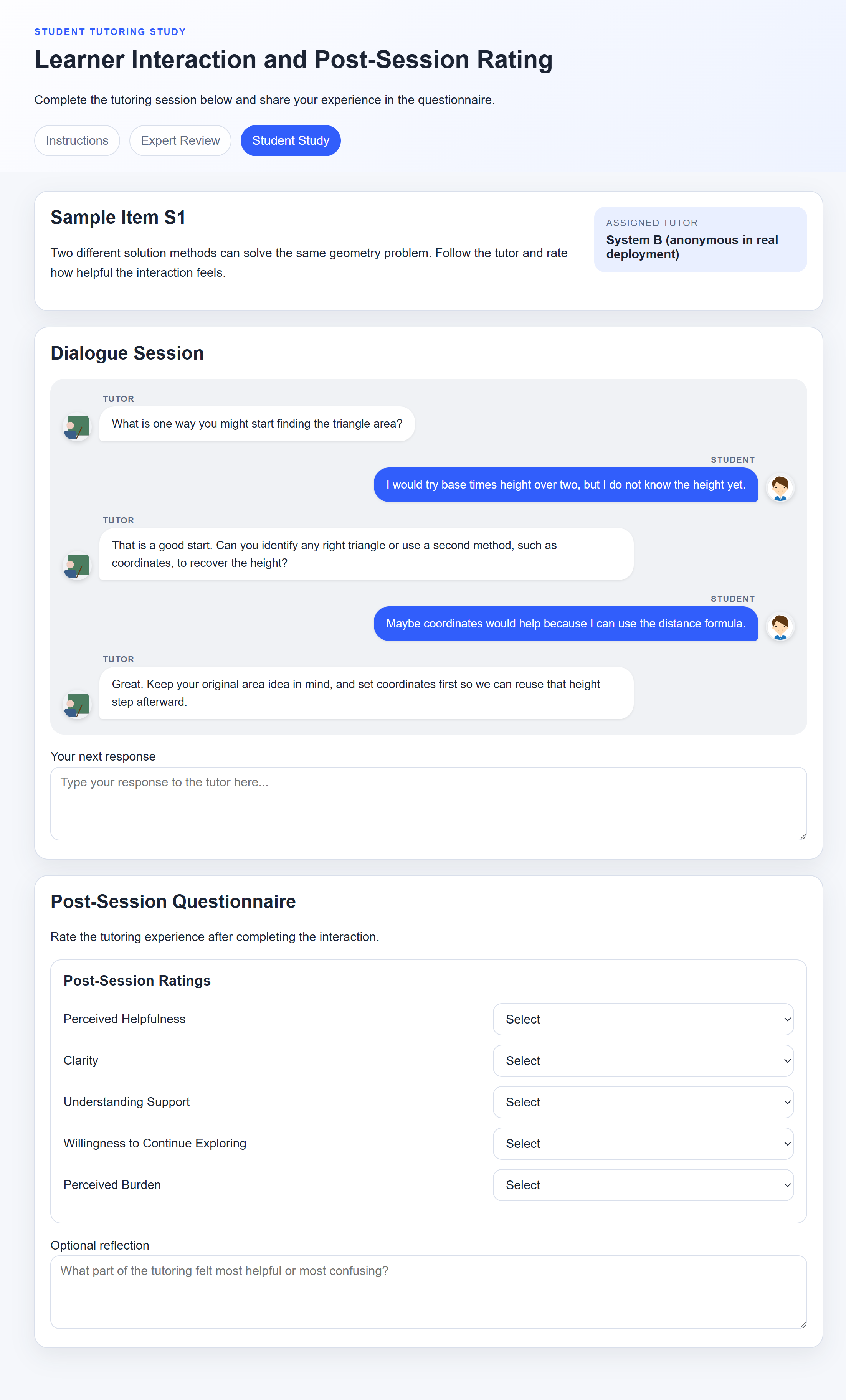}
\caption{\textbf{Student-study interface.} The learner-facing page displays the tutoring dialogue in a chat format and collects five-dimension post-session Likert ratings with optional free-text reflection.}
\label{fig:student_interface_placeholder}
\end{figure*}

Figure~\ref{fig:expert_interface_placeholder} shows the blinded expert-review interface presented to teacher and teaching-assistant annotators, where anonymized tutoring dialogues from two systems are displayed side-by-side for Likert scoring.
Figure~\ref{fig:student_interface_placeholder} shows the learner-facing interface used in the pilot student study, where participants interact with an anonymized tutoring system via a chat-style window before completing the post-session questionnaire.
Both interfaces are designed to conceal system identities throughout the evaluation session.

\subsection{Detailed Evaluation Results}

Table~\ref{tab:student_eval} in the main text reports the Track A learner self-report results across five subjective dimensions.
Here, Table~\ref{tab:expert_eval} provides the complementary blinded teacher/expert evaluation, and Table~\ref{tab:llm_judge_eval} presents Track B independent LLM judge scores along with human--LLM rank-correlation with teacher and student ratings.
Taken together, these results show that ToST consistently achieves the highest or near-highest ratings across human participants and the independent LLM judge, with strong agreement between human raters and the LLM judge ($\rho_T, \rho_S > 0.7$ for all systems).

The expert ratings further qualify the benchmark results.
ToST is not uniformly best on every teacher-rated dimension: DeepSeek V3.2 is marginally higher on diagnostic accuracy and cognitive load appropriateness.
For cognitive load, this likely reflects the tradeoff between directness and structured exploration.
DeepSeek V3.2 tends to give more linear, answer-oriented guidance, which can feel lower-friction to expert raters because the dialogue asks the learner to make fewer path-management decisions.
ToST explicitly maintains multi-path state, preserves correct intermediate reasoning, and guides path transitions; these operations introduce more instructional structure and active reasoning demands, so they can be judged slightly less lightweight even when pedagogically useful.
However, ToST receives the strongest ratings on guidance naturalness, preservation of correct intermediate reasoning, switch naturalness, and overall helpfulness, which are the dimensions most directly tied to multi-path tutoring control.

\section{Parser Agreement Analysis of Parallel Reasoning Tree Parsing}
\label{sec:appendix_parse}

As the PRT forms the foundation of our framework, we evaluate its consistency by cross-validating the parsing process.
We sampled one-eighth of the student responses (332 samples) from the MPSG-Bench test set and independently re-parse them using DeepSeek V3.2, GPT-5, and Gemini-2.5-pro with an identical prompt (as shown in Appendix~\ref{sec:appendix_prompt}).
Consistency is measured via Path Diversity Consistency (PDC; Jaccard similarity over paths) and Node Match Rate (NMR; embedding-based semantic alignment).
As shown in Table~\ref{tab:parse}, average PDC and NMR reach 99.23\% and 97.69\%, respectively, indicating high parser agreement across model choices. We emphasize that agreement does not guarantee semantic correctness; rather, it suggests relative stability of the parsing pipeline under alternative LLM backends.

\end{document}